%% file: main.tex
\documentclass[10pt,twocolumn,letterpaper]{article}
\usepackage{cvpr}
\usepackage{placeins}
\input{preamble}

\input{macros}

\usepackage[normalem]{ulem}

\let\cvprparagraph\paragraph
\renewcommand{\paragraph}[1]{\vspace{.5em}\noindent\textbf{#1.}}
\definecolor{cvprblue}{rgb}{0.21,0.49,0.74}
\usepackage[pagebackref,breaklinks,colorlinks,allcolors=cvprblue]{hyperref}
\AtBeginDocument{%
  \crefname{appendix}{Appendix}{Appendices}%
  \Crefname{appendix}{Appendix}{Appendices}%
}
\def\confName{CVPR}
\def\confYear{2027}

\title{Boundary-Aware Policy Learning for Agent-Orchestrated Robot Manipulation}
\hypersetup{pdftitle={\thetitle}}
\input{arxiv_authors}
\author{\arxivauthors}
\newcommand{\mref}[1]{\cref{#1}}
\makeatletter
\apptocmd{\@maketitle}{%
  \begingroup\input{fig/teaser}\par\endgroup
\vspace{12pt}}{}%
  {\PackageError{bapl}{Could not add the teaser to the title block}{}}
\makeatother

\begin{document}
\maketitle
\arxivwebsitefootnote
\begingroup
  \brokenpenalty=10000 %
  \input{sections/0_abstract}\par
\endgroup
\input{sections/1_introduction}

\input{sections/2_related}
\input{sections/3_method}
\input{sections/4_experiments}

\input{sections/5_conclusion}
\input{sections/6_llm_usage_statement}
{
    \small
    \bibliographystyle{ieeenat_fullname}
    \bibliography{references,references_supp}
}

\clearpage
\let\paragraph\cvprparagraph
\setlength{\textfloatsep}{12pt plus 2pt minus 2pt}
\setlength{\dbltextfloatsep}{12pt plus 2pt minus 2pt}
\renewcommand{\thefigure}{S\arabic{figure}}
\renewcommand{\thetable}{S\arabic{table}}
\setcounter{figure}{0}
\setcounter{table}{0}
\renewcommand{\theHfigure}{supp.\arabic{figure}}
\renewcommand{\theHtable}{supp.\arabic{table}}
\input{sections/A_appendix}
\end{document}

%% file: preamble.tex
\usepackage{cuted}    %
\usepackage{currfile}
\usepackage{caption}  %

\input{figures_preamble}

\usetikzlibrary{decorations.pathmorphing}
\usepackage[most]{tcolorbox}
\usepackage{longtable}
\usepackage{tabularx}
\usepackage{colortbl}
\usepackage{pifont}
\definecolor{oursrow}{rgb}{0.88,1.00,1.00}

\newcommand{\oursL}[1]{\multicolumn{1}{@{}>{\columncolor{oursrow}[0pt][\tabcolsep]}l}{#1}}
\newcommand{\oursR}[1]{\multicolumn{1}{>{\columncolor{oursrow}[\tabcolsep][0pt]}c@{}}{#1}}

%% file: figures_preamble.tex
\usepackage{tikz}
\usetikzlibrary{positioning,arrows.meta,fit,calc,backgrounds,decorations.pathreplacing,shapes.geometric}

\definecolor{bpOurs}{RGB}{0,114,178}      %
\definecolor{bpPrior}{RGB}{213,94,0}      %
\definecolor{bpData}{RGB}{0,158,115}      %
\definecolor{bpAccent}{RGB}{230,159,0}    %
\definecolor{bpGray}{RGB}{110,110,110}    %
\definecolor{bpLight}{RGB}{238,238,238}   %
\colorlet{bpOursFill}{bpOurs!18}
\colorlet{bpPriorFill}{bpPrior!15}
\colorlet{bpDataFill}{bpData!15}

\tikzset{
  bp/base/.style   ={draw=black!70, line width=0.5pt, rounded corners=2pt, align=center,
                     font=\sffamily\scriptsize, minimum height=0.6cm, inner sep=3pt},
  bp/module/.style ={bp/base, fill=white},
  bp/frozen/.style ={bp/base, fill=bpLight, draw=bpGray, text=black!80},
  bp/ours/.style   ={bp/base, fill=bpOursFill, draw=bpOurs, line width=0.7pt},
  bp/prior/.style  ={bp/base, fill=bpPriorFill, draw=bpPrior},
  bp/data/.style   ={bp/base, fill=bpDataFill, draw=bpData},
  bp/todo/.style   ={bp/base, dashed, draw=bpGray, text=bpGray},
  bp/flow/.style   ={-{Latex[length=1.6mm]}, line width=0.6pt, black!75},
  bp/note/.style   ={font=\sffamily\tiny, text=bpGray, align=center},
  bp/panel/.style  ={font=\sffamily\footnotesize\bfseries, anchor=west},
  bp/frame/.style  ={draw=black!70, fill=white, line width=0.4pt},
}

%% file: macros.tex
\usepackage{multirow}
\usepackage{subcaption}

\newcommand{\tighten}{\looseness=-1 }
\newcommand{\appref}[1]{\mbox{\textbf{\cref{#1}}}}

%% file: arxiv_authors.tex
\newcommand{\arxivauthors}{%
  Kaipeng Fang\textsuperscript{1,2,}\thanks{Equal contribution.\quad
    \textsuperscript{\ensuremath{\dagger}}Corresponding authors.}\quad
  Yuhao Wang\textsuperscript{1,*}\quad
  Ji Zhang\textsuperscript{1}\quad
  Yunsong Zhou\textsuperscript{2,\ensuremath{\dagger}}\quad
  Jingkuan Song\textsuperscript{3,4}\quad
  Lianli Gao\textsuperscript{1,\ensuremath{\dagger}}\\[0.4em]
  {\small \textsuperscript{1}University of Electronic Science and Technology of China}\\
  {\small
    \textsuperscript{2}Shanghai Artificial Intelligence Laboratory\quad
    \textsuperscript{3}Tongji University\quad
    \textsuperscript{4}Shanghai Innovation Institute}%
}
\newcommand{\arxivwebsitefootnote}{%
  \begingroup
  \renewcommand{\thefootnote}{}%
  \urlstyle{same}%
  \footnotetext[0]{Project website: \href{https://kaipengfang.github.io/bapl-agentic-manip}{\nolinkurl{kaipengfang.github.io/bapl-agentic-manip}}}%
  \endgroup
}
\hypersetup{pdfauthor={Kaipeng Fang, Yuhao Wang, Ji Zhang, Yunsong Zhou, Jingkuan Song, Lianli Gao}}

%% file: fig/teaser.tex
\centering
\includegraphics[width=0.925\linewidth]{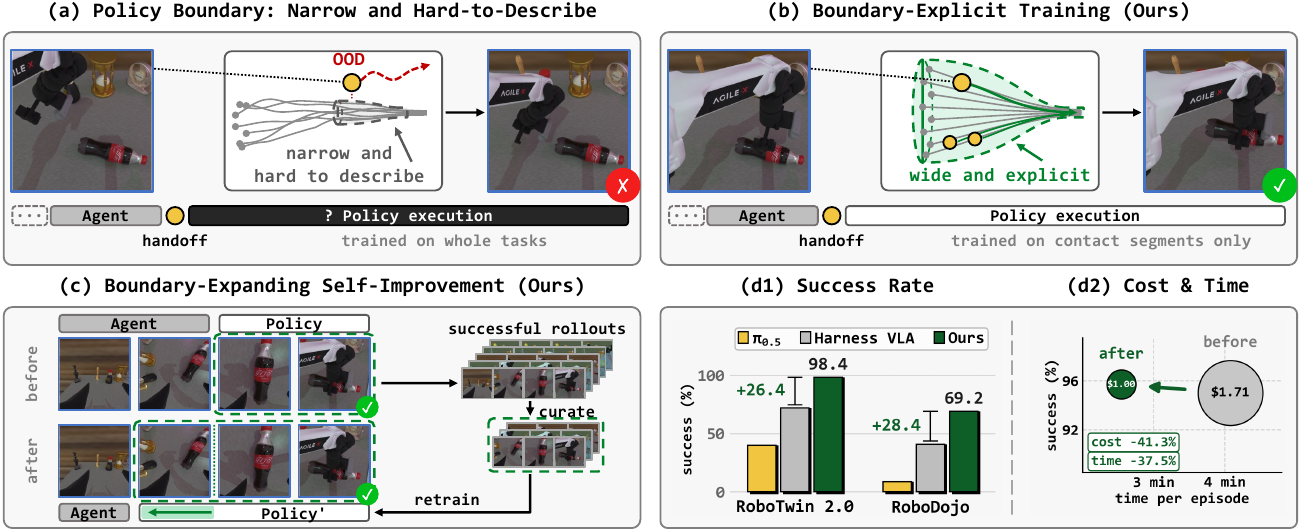}
\captionsetup{hypcap=false}
\captionof{figure}{\textbf{Motivation and Method Overview.} (a) A policy trained on whole tasks often has a narrow and hard-to-describe handoff boundary, making agent handoffs prone to fall outside the policy's reliable operating region (OOD). (b) \textbf{Boundary-Explicit Training} trains the policy on contact segments with diverse start poses, producing a wider and explicit handoff boundary that enables reliable agent--policy coordination. (c) \textbf{Boundary-Expanding Self-Improvement} retrains the policy on curated successful system rollouts (Policy$'$), extending the boundary toward earlier task stages and allowing the agent to hand over sooner. (d) Our method substantially improves task success in randomized scenes of RoboTwin 2.0 and on RoboDojo (d1), while reducing API cost (bubble size) and execution time (d2).}
\captionsetup{hypcap=true}
\label{fig:teaser}

%% file: sections/0_abstract.tex
\begin{abstract}
Agent-orchestrated robot systems combine the general reasoning capability of multimodal agents with the precise physical control of learned visuomotor policies. Yet such systems often fail at the moment of handoff: a policy is reliable only from a limited set of start states, while the agent may invoke it outside this region. We call this region the policy's handoff boundary. Existing whole-task policies typically have handoff boundaries that are narrow and difficult for an agent to characterize, making reliable coordination challenging. We propose \textbf{Boundary-Aware Policy Learning}, which explicitly shapes and then expands the handoff boundary of a visuomotor policy. First, \textbf{Boundary-Explicit Training} fine-tunes the policy on contact segments with diverse start poses, yielding a wide and well-defined boundary that can be communicated to the agent through a simple handoff protocol. Second, \textbf{Boundary-Expanding Self-Improvement} retrains the policy on curated successful system rollouts, extending the boundary toward earlier stages of each task and allowing the agent to hand over sooner with fewer language-model calls. Across RoboTwin 2.0 and RoboDojo, our system outperforms the strongest baseline by over \textbf{26} percentage points, reaching 98.4\% success in randomized RoboTwin 2.0 scenes, while further reducing its API cost and execution time by \textbf{41.3}\% and \textbf{37.5}\%, respectively. These results suggest that the key to robust agent--policy orchestration lies not only in improving either component individually, but also in explicitly shaping the \mbox{boundary between them.}
\end{abstract}

%% file: sections/1_introduction.tex
\section{Introduction}
\label{sec:intro}

\tighten Robot manipulation needs both task-level decisions and precise physical interaction with objects. Agents built on general-purpose multimodal models now handle the task level through tool calls: they capture images, segment objects, plan multi-step tasks and recover when a step fails~\cite{codeaspolicies2023,voxposer2023,maestro2025,alrm2026,roboclaw2026}. Such an agent generalizes across layouts and instructions but executes contact poorly, whereas a learned visuomotor policy executes contact well. Recent systems therefore combine their strengths: the agent orchestrates the task and hands over physical interaction with objects to such a policy~\cite{harnessvla2026,huang2026roboharness,xu2026baton,semantichandoff2026,robocop2026,hu2026orchestrating}, a paradigm we refer to as \emph{agent-orchestrated robot manipulation}.

\looseness=1 However, a policy completes its command reliably only from a limited range of states, and the agent may hand over outside this range. We refer to this range as the \emph{handoff boundary} of the policy. These systems call an off-the-shelf policy that is trained to complete whole tasks from initial states, which leaves this boundary narrow and hard to describe in the middle of a task (\cref{fig:teaser}a). As a result, the agent tends to hand over the whole task from the initial pose. When the policy fails, the agent corrects the arm and hands over once more, but often from a state outside the boundary, so the policy fails again, as we observe in \cref{sec:boundary}. Existing systems mostly mitigate such failures on the agent side, through exploration, memory and verification~\cite{harnessvla2026,huang2026roboharness,semantichandoff2026,xu2026baton}, but these measures cannot change the boundary, which is mainly set by the data on which the policy is trained. This raises a natural question: \emph{Can a policy be trained to have a handoff boundary that is wide and explicit to the agent, for more \mbox{reliable agent--policy coordination?}}

\tighten In this paper, we propose \textbf{Boundary-Aware Policy Learning}. The key idea is to define and expand the handoff boundary through training, instead of constraining the agent through prompts or letting the agent discover the boundary through exploration. In its first stage, \textbf{Boundary-Explicit Training}, the policy is fine-tuned only on contact segments, in which the robot physically interacts with objects. We collect these segments with diverse start poses around the target object to make the boundary wide (\cref{fig:teaser}b), and the well-defined rules for collecting them make the boundary explicit. With a short handoff protocol that describes this boundary, the agent calls the policy reliably.

However, these reliable calls come at a cost: the policy covers only the contact segment, and the agent completes the rest of each task through many slow and costly language-model calls. To reduce these calls, the second stage, \textbf{Boundary-Expanding Self-Improvement}, first collects the successful rollouts of entire tasks that the system produces and removes their suboptimal segments with a data curation pipeline, as illustrated in \cref{fig:teaser}c. We then retrain the policy on these curated rollouts together with the contact segments from the first stage, so that the handoff boundary extends to earlier stages of each task while preserving the boundary learned in the first stage. After re-exploring each task with the retrained policy, the agent hands over earlier and needs fewer language-model calls. In summary, our contributions are as follows:

\begin{itemize}
\item We propose Boundary-Aware Policy Learning, which defines and expands the handoff boundary through training, for more reliable agent--policy coordination.
\item To this end, we introduce Boundary-Explicit Training (\cref{sec:aligned_learning}), which makes the handoff boundary wide and explicit, and Boundary-Expanding Self-Improvement (\cref{sec:expand}), which extends it to earlier stages of each task.
\item With boundary-explicit training alone, our system outperforms the strongest baseline by more than \textbf{26} percentage points in randomized scenes of RoboTwin~2.0~\cite{robotwin2_2025} and on RoboDojo~\cite{chen2026robodojo}. On RoboTwin~2.0, self-improvement further reduces its API cost and time per episode by \textbf{41.3\%} and \textbf{37.5\%} on average, without loss in success rate (\cref{fig:teaser}d).
\end{itemize}

%% file: sections/2_related.tex
\section{Related Work}
\label{sec:related}

\paragraph{Agentic robot manipulation}
\tighten Language models were first used as robot planners that compose a fixed set of primitives~\cite{saycan2022,codeaspolicies2023,innermonologue2022,voxposer2023}.
More recently, frontier models have made it possible for an agent to control a robot arm directly, with no robot-specific training~\cite{astranomy2026,sensenova2026astra,astra2026robodojo,ora2026}.
However, such an agent still falls short on physical interaction with objects, so a growing line of work on agent-orchestrated robot manipulation integrates a learned visuomotor policy, typically a vision-language-action (VLA) model, into the agent framework~\cite{harnessvla2026,chen2026volo,huang2026roboharness,galanti2026gap,zhao2026retriever,xu2026baton,claro2026,maestro2025}.
In these systems, the agent decides when to call the policy and how to instruct the policy, and the interface between the agent and the policy becomes the main bottleneck~\cite{hu2026orchestrating}.
Most existing attempts address this bottleneck from the agent side by adding verification and memory around a frozen policy~\cite{semantichandoff2026,huang2026roboharness,xu2026baton}.
In this paper, we address this bottleneck from the policy side: we train the policy so that its handoff boundary is wide and explicit to the agent, and then expand this boundary as the system runs.

\paragraph{Robot skill learning}
Rather than learning one policy for the whole task, a long line of work decomposes tasks into reusable skills and composes them at run time~\cite{robomatrix2024,atomiclib2025,voyager2023,aspire2026,li2026recover}.
With generalist VLA policies~\cite{rt2_2023,openvla2024,octo2024,pi0_2024,pi05_2025}, these skills are now fine-tuned from, or prompted out of, one shared model~\cite{primskills2026,atomicvla2026,mastvla2026,skillvla2026,skillretrieval2026}.
Each such skill can start only from certain states, which are classically modeled as the initiation set of an option~\cite{sutton1999options} and are learned so that skills can be chained~\cite{konidaris2009skillchaining}. In this paper, the contact segments on which we fine-tune the policy in boundary-explicit training can also be viewed as such skills, but we do so to give the policy an explicit handoff boundary, so that the agent knows when and where to hand over.

\paragraph{Self-improving robot systems}
Beyond a fixed memory, a robot system can keep improving after deployment.
Early work rewrites the agent's memory from failures~\cite{reflexion2023,voyager2023}.
More recently, robot agents discover new skills~\cite{aspire2026,li2026recover}, curate their own executions into training data~\cite{chen2025curating,roboclaw2026,enpire2026}, and fine-tune the policy from model-guided exploration~\cite{eximo2026,insight2026,agenticvla2026}.
Closest to our work, Robo-COP fine-tunes a VLA policy on its own executions to fix recurring failures and revises the agent's memory after each update~\cite{robocop2026}.
In this paper, we retrain the policy on the system's successful rollouts and let the agent re-explore each task with the retrained policy, so that the agent hands over earlier and needs \mbox{fewer language-model calls.}

%% file: sections/3_method.tex
\input{fig/paradigm}
\input{fig/overview}
\section{Method}
\label{sec:method}

In this paper, we study agent-orchestrated robot manipulation, which combines a general-purpose agent, equipped with a system prompt, a memory and tools, with a learned visuomotor policy to solve robot manipulation tasks, as shown in \cref{fig:paradigm}. We consider this paradigm a better choice than letting the agent control the robot directly, which we examine in an early exploratory study (\appref{sec:supp_closer_look}). Boundary-Aware Policy Learning improves this system through the handoff between the agent and the policy: boundary-explicit training (\cref{sec:aligned_learning}) makes the handoff reliable, and boundary-expanding self-improvement (\cref{sec:expand}) lets the agent hand over earlier.

\subsection{Preliminaries}
\label{sec:prelim}

\paragraph{Task setup}
In a robot manipulation task, a robot acts in an environment to follow a language instruction $\ell$. At each control step $t$, the robot provides an observation $o_t$ with RGB images, depth maps and the proprioceptive state, and a policy $\Pi$ outputs an action $a_t\sim\Pi(\cdot\mid o_t,\ell)$. The robot executes $a_t$, which changes the environment, and this loop repeats until the task is completed.

\paragraph{Agent-orchestrated robot manipulation}
Here, we use a general-purpose agent $\mathcal{A}$ that orchestrates a learned visuomotor policy $\pi_\theta$ to solve the manipulation task. The agent calls tools in a loop, where the $k$-th call $c_k\sim\mathcal{A}(\cdot\mid\ell,M,h_k)$ is conditioned on the instruction, a memory $M$ and the history $h_k$ of earlier calls. Each call either perceives the scene, reads or writes the memory, or controls the robot through motion tools, such as a planner that reaches an end-effector target $x$, or through the policy $\pi_\theta$, which runs with a short command $u$ for $n$ chunks of $H$ actions. After each call, the agent receives the new observation, and this loop continues until the agent finishes the task or a budget of calls is spent. Following Harness VLA~\cite{harnessvla2026}, the agent first explores each task and writes what it learns into the memory $M$, which stays fixed at evaluation.

\paragraph{Handoff boundary}
\tighten When the agent calls the policy, it hands over control of the robot to the policy, and we refer to this transfer as a handoff. A handoff is reliable only if the agent hands over in a state from which the policy can complete its command. Formally, we define the handoff boundary $\mathcal{B}_\epsilon(\pi_\theta,u)$ as the set of observations from which the policy completes command $u$ with probability at least $1-\epsilon$:
\begin{equation}
  \mathcal{B}_\epsilon(\pi_\theta,u)=\left\{\, o \;:\; P_\theta(o,u) \ge 1-\epsilon \,\right\},
  \label{eq:boundary}
\end{equation}
where $P_\theta(o,u)$ is the probability that $\pi_\theta$ completes command $u$ when it starts from observation $o$. Since this probability cannot be measured for every observation, we estimate the boundary in \cref{sec:boundary} from the success rate of the policy at increasing distances from its training trajectories.

\subsection{Boundary-explicit training}
\label{sec:aligned_learning}

\label{sec:aligned}
The handoff boundary $\mathcal{B}_\epsilon(\pi_\theta,u)$ in \cref{eq:boundary} is mainly set by the data on which the policy is trained. A whole-task policy, which is trained on complete demonstrations of entire tasks, often has a handoff boundary that is narrow and hard to describe in the middle of a task.

\paragraph{Data collection and policy training}
To address this problem, we first define the roles of the agent and the policy: the policy handles only the contact segments of each task, in which the robot physically interacts with objects, and the agent does the rest. The handoff boundary therefore lies near the graspable points of the target object. We then collect demonstrations of contact segments from diverse start poses within this region to widen the boundary, as shown in the data-collection scene of \cref{fig:overview}a. Each demonstration in the resulting object-interaction data $\mathcal{D}_0$ pairs a command $u$, such as ``Pick up the stapler with the left arm'', with a sequence of observation--action pairs $(o_t,a_t)$. For example, on RoboTwin~2.0, we collect demonstrations of 147 subtasks across the 50 tasks, and further details of $\mathcal{D}_0$ are provided in \appref{sec:supp_stage1}.
We fine-tune a pretrained policy, $\pi_{0.5}$~\cite{pi05_2025}, on $\mathcal{D}_0$ with end-effector poses as the action representation, which matches the end-effector targets that the agent specifies.
The resulting policy thus has a wide and explicit handoff boundary.

\paragraph{Handoff protocol}
\label{sec:protocol}
The handoff protocol describes this boundary to the agent as rules in its prompt, which restate how the start poses of a grasp were sampled (8 to 12\,cm above the grasp point) and how to hand over within the boundary. Since the boundary is explicit, these rules need no example images or task-specific descriptions, and the same protocol applies to every task. Under these rules, the agent calls the policy only for a contact segment, with a short command $u$ that names the segment. Before each call, the agent locates the target object with its perception tools and moves the gripper into the boundary with one planner motion. The agent then runs the policy for at most four chunks, so that the call stays short enough to check. After the call, the agent checks the result in fresh observations and takes over again. If the contact segment has failed, the agent observes the scene again and moves the gripper back into the boundary before the next call. The bottom of \cref{fig:overview}a shows one such call, from finding the object to the check.

\subsection{Boundary-expanding self-improvement}
\label{sec:expand}

With boundary-explicit training, the agent hands over only the contact segments and completes the rest of each task through many slow and costly language-model calls. Boundary-expanding self-improvement reduces these calls by expanding the handoff boundary toward earlier stages of each task, so that the agent can hand over earlier. The training data for this expansion come from the system itself, since its successful rollouts of entire tasks match its own distribution. However, these rollouts should not be imitated directly, because they contain suboptimal segments, such as repeated adjustments of the arm pose and retries after a failed policy call.

\paragraph{Data curation}
To remove these segments, we use a data curation pipeline of two steps. A curation agent first labels every segment of a rollout as useful or suboptimal and keeps the useful segments as a reference. Since directly removing the suboptimal segments would leave abrupt jumps in the robot state between the remaining segments, we then let the agent $\mathcal{A}$ re-execute the task in the same scene following this reference and record the re-executed rollouts as the final system rollout data $\mathcal{D}_1$, as shown at the top of \cref{fig:overview}b.

\paragraph{Retraining and re-exploration}
\tighten We retrain the policy on $\mathcal{D}_0$ and $\mathcal{D}_1$ together, and the retrained parameters $\theta'$ minimize its training loss $\mathcal{L}$ over both datasets:
\begin{equation}
  \theta' = \arg\min_{\theta}\; \mathbb{E}_{(o_t,a_t,g)\sim\mathcal{D}_0\cup\mathcal{D}_1}\, \mathcal{L}\big(\pi_\theta(\cdot\mid o_t,g),\,a_t\big),
  \label{eq:retrain}
\end{equation}
where the language input $g$ is the command $u$ of a contact segment in $\mathcal{D}_0$ or the instruction $\ell$ of a whole task in $\mathcal{D}_1$. The policy thus learns to complete a whole task from the instruction while keeping its ability to perform a single contact segment from a command.
The whole-task rollouts in $\mathcal{D}_1$ extend the handoff boundary to earlier stages of each task, so that the agent can hand over before the gripper reaches the target object. Since $\mathcal{D}_1$ records each task in the way the agent orchestrates it, its contact segments also start inside the boundary learned in \cref{sec:aligned_learning} and thus agree with $\mathcal{D}_0$, so the agent can still hand over in the middle of a task. To adapt to the retrained policy, we update the handoff protocol to describe the expanded boundary, and the agent re-explores each task and updates its memory $M$, which again stays fixed at evaluation. Under the updated protocol, the agent first hands over the whole task from the initial pose and falls back to contact-segment calls when the progress stalls or a step fails. As a result, the agent needs fewer language-model calls, as the bottom of \cref{fig:overview}b illustrates.

%% file: fig/paradigm.tex
\begin{figure}[t]
\centering
\includegraphics[width=\linewidth]{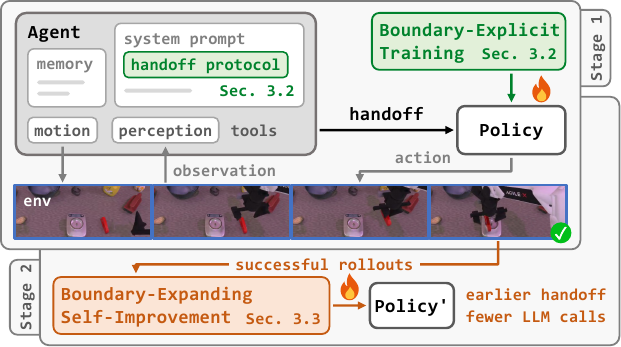}
\caption{\textbf{Illustration of agent-orchestrated robot manipulation, with the components of our method highlighted in color.}}
\label{fig:paradigm}
\end{figure}

%% file: fig/overview.tex
\begin{figure*}[t]
\centering
\includegraphics[width=0.980\linewidth]{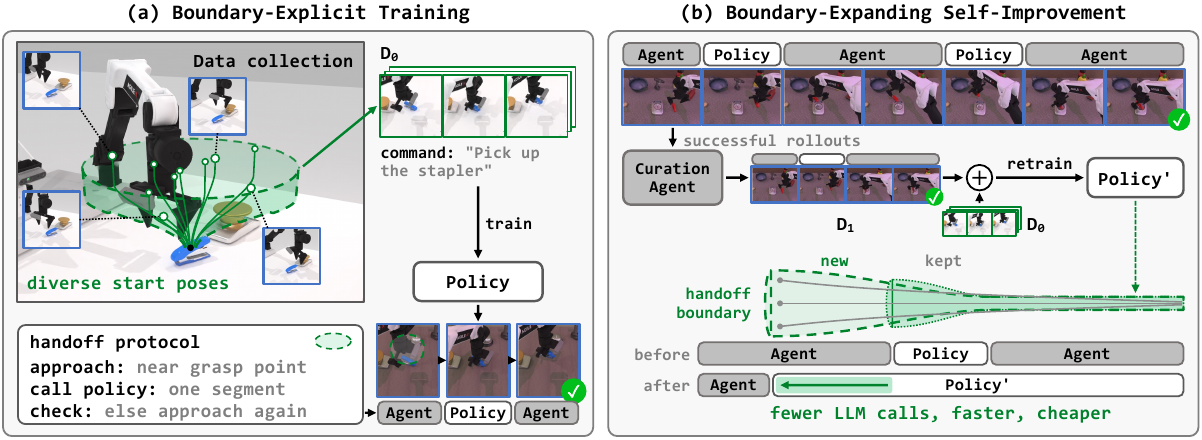}%
\caption{\textbf{Illustration of Boundary-Aware Policy Learning.} (a) \textbf{Boundary-Explicit Training} collects contact-segment demonstrations $\mathcal{D}_0$ from diverse start poses near the target object and fine-tunes the policy on these segments. A simple handoff protocol then guides the agent to enter the policy's reliable region, invoke it for one contact segment, and verify the outcome. (b) \textbf{Boundary-Expanding Self-Improvement} collects successful system rollouts, removes suboptimal behaviors through curation, and re-executes the tasks to construct $\mathcal{D}_1$. Retraining the policy on $\mathcal{D}_0 \cup \mathcal{D}_1$ expands its handoff boundary toward earlier task stages, enabling earlier handoffs \mbox{with fewer language-model calls.}}
\label{fig:overview}
\vspace{-4pt}%
\end{figure*}

%% file: sections/4_experiments.tex
\section{Experiments}
\label{sec:exp}

\subsection{Setup}
\label{sec:setup}

\paragraph{Benchmarks and protocol}
We evaluate the system after boundary-explicit training on RoboTwin 2.0~\cite{robotwin2_2025} and RoboDojo~\cite{chen2026robodojo}. On RoboTwin 2.0, we collect the object-interaction data $\mathcal{D}_0$ with the benchmark's scripted expert and conduct agent exploration in clean scenes, then evaluate each task over 10 episodes in each of two settings: in clean scenes (C2C) and under the official randomized configuration (C2R). On RoboDojo, we evaluate on the 26 tasks in the generalization, memory and open-vocabulary subsets, with 10 episodes per task. For generalization tasks, we collect $\mathcal{D}_0$ and conduct agent exploration in the clean variant, then evaluate in the randomized variant, as in C2R. For memory and open-vocabulary tasks, data collection and agent exploration use seeds disjoint from evaluation.

For self-improvement evaluation, we use RoboTwin 2.0, where both the collection of the system rollout data $\mathcal{D}_1$ and agent re-exploration take place in randomized scenes. Since this randomized configuration is now used during self-improvement, we define two evaluation settings to assess generalization. The easy setting uses unseen seeds of the same randomized configuration. The hard setting also uses unseen seeds, replaces the textures and the distractor objects with assets imported from RoboDojo and changes the lighting, none of which appear during training or exploration. In both settings, we report the success rate together with the number of language-model calls, the API cost and the time per episode.

\paragraph{Baselines}
\looseness=1 We compare our method with representative baselines for end-to-end policy execution and agent-based robot control. As an end-to-end learned policy baseline, we fine-tune $\pi_{0.5}$~\cite{pi05_2025} on the official whole-task demonstrations in clean scenes together with $\mathcal{D}_0$, so that it learns from the same contact segments as our policy, and evaluate it end to end. AGP~\cite{jia2026agp} represents agents with runtime programming, in which the agent writes and revises programs for perception and robot control during execution. ORA~\cite{ora2026} represents agents with predefined perception and motion tools, and neither AGP nor ORA uses a learned policy. Harness VLA~\cite{harnessvla2026} represents agents that orchestrate a learned policy. Following its official setting, we equip it with LingBot-VLA~\cite{lingbotvla2026}, a whole-task policy fine-tuned on RoboTwin 2.0 and released with Harness VLA. On RoboDojo, we again follow its official setting and equip it with VPP2~\cite{guo2026vpp2}. For all agent-based methods that we run, including ours, we use Claude Opus 5.5 with high reasoning effort. \appref{sec:supp_setup,sec:supp_stage1,sec:supp_prompt} detail the evaluation protocol and the tools of the agent, the training of our policy and \mbox{the prompt of the agent.}

\subsection{Main results}
\label{sec:main}

\paragraph{An explicit handoff boundary brings large gains on both benchmarks}
We compare the system after boundary-explicit training with the baselines in \cref{tab:main}, which leads to two conclusions. \textbf{First}, an explicit handoff boundary brings large gains even over Harness VLA, which also orchestrates a learned policy: our system is 26.4 percentage points higher on RoboTwin 2.0 under the official randomized configuration (C2R) and 28.4 percentage points higher on average on RoboDojo. \textbf{Second}, handing over only the contact segments makes the system less sensitive to scene randomization. A whole-task policy can complete most tasks in clean scenes (C2C), where $\pi_{0.5}$ alone reaches 85.8\% and Harness VLA reaches 96.0\%, but randomization lowers them to 40.0\% and 72.0\%, whereas the success rate of our system does not drop.

\paragraph{The agent and the policy complement each other}
Taking a closer look at the subsets of RoboDojo, we find two interesting observations. \textbf{First}, memory and open-vocabulary tasks rely on the agent: the whole-task policy $\pi_{0.5}$ rarely succeeds, whereas AGP, an agent without a learned policy, performs much better on both. \textbf{Second}, AGP achieves only 10.0\% on the generalization subset, since its tasks test generalization to new scenes through diverse physical interactions with objects. By combining the two, our system leads in all three subsets, with the agent handling task-level reasoning and orchestration and the policy providing closed-loop control for physical interaction with objects.

\begin{table}[t]
\centering
\caption{\textbf{Success rates (\%) on RoboTwin 2.0 and RoboDojo.} On RoboTwin 2.0, data collection and exploration use clean scenes, and evaluation uses clean (C2C) or randomized (C2R) scenes. Gen., Mem.\ and Open denote the generalization, memory and open-vocabulary subsets. Shaded columns are the main results: the harder C2R setting and the task-weighted average (Avg.). $^\dagger$Reported by the official release (GPT-6 Astra, \mbox{low reasoning effort).}}
\label{tab:main}
\vspace{2pt}
\small
\setlength{\tabcolsep}{1.8pt}
\renewcommand{\arraystretch}{1.0}%
\colorlet{keycol}{oursrow}%
\newcommand{\deepstrut}{\rule[-\dimexpr\dp\strutbox+\aboverulesep\relax]{0pt}{0pt}}%
\newcommand{\highstrut}{\rule{0pt}{\dimexpr\ht\strutbox+\belowrulesep\relax}}%
\begin{tabularx}{\columnwidth}{@{}l >{\centering\arraybackslash}X >{\columncolor{keycol}\centering\arraybackslash}X *{3}{>{\centering\arraybackslash}X} >{\columncolor{keycol}[\tabcolsep][0pt]\centering\arraybackslash}X@{}}
\toprule
\multirow{2}{*}{\textbf{Method}} & \multicolumn{2}{c}{\textbf{RoboTwin 2.0}} & \multicolumn{4}{c}{\textbf{RoboDojo}} \\
\cmidrule(lr){2-3}\cmidrule(lr){4-7}
& C2C & \textbf{C2R}\deepstrut & Gen. & Mem. & Open & \textbf{Avg.} \\
\specialrule{\lightrulewidth}{0pt}{0pt}
$\pi_{0.5}$~\cite{pi05_2025}\highstrut & 85.8 & 40.0 & 16.7 & 3.3 & 0.0 & 8.5 \\
AGP~\cite{jia2026agp} & 54.0 & 56.0 & 10.0 & 46.7 & 50.0 & 30.8 \\
ORA~\cite{ora2026} & 74.0 & 70.0 & -- & -- & -- & -- \\
Harness VLA~\cite{harnessvla2026} & \textbf{96.0} & 72.0 & 41.7$^\dagger$ & 36.7 & 42.5 & 40.8 \\
\textbf{Ours}\deepstrut & 95.6 & \textbf{98.4} & \textbf{62.5} & \textbf{80.0} & \textbf{71.3} & \textbf{69.2} \\
\specialrule{\heavyrulewidth}{0pt}{\belowbottomsep}
\end{tabularx}
\vspace{0pt}%
\end{table}

\begin{figure*}[t]
\centering
\includegraphics[width=\textwidth]{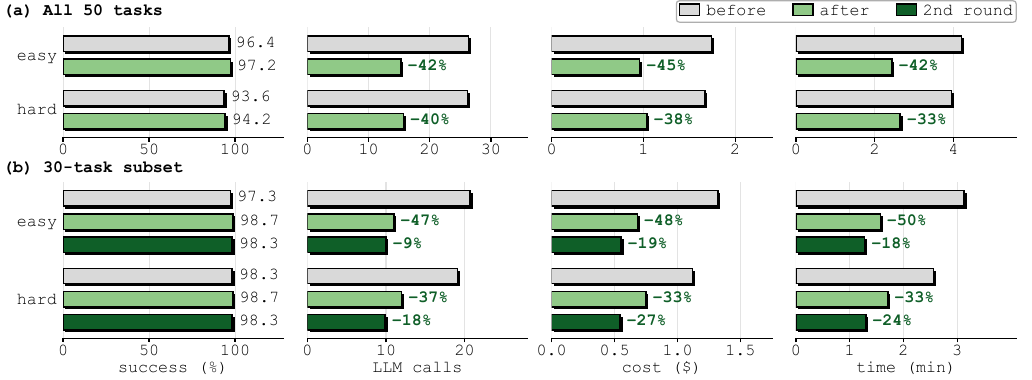}
\caption{\textbf{Self-improvement makes the system faster and cheaper to run.} Success rate, language-model calls, API cost and time per episode of our system on RoboTwin 2.0 in the easy and hard settings, with 10 episodes per task. (a) Before and after self-improvement on all 50 tasks. (b) A second round on 30 randomly selected tasks. Percentages are relative to the previous stage.}
\label{fig:selfimprove}
\vspace{-4pt}%
\end{figure*}

\subsection{Self-improvement results}
\label{sec:expandexp}

\paragraph{Self-improvement makes the system faster and cheaper to run}
To evaluate boundary-expanding self-improvement, we compare the system before and after this stage on all 50 tasks of RoboTwin 2.0 in the easy and hard settings. As shown in \cref{fig:selfimprove}a, the API cost and time per episode at execution fall by 41.3\% and 37.5\% on average, while the success rate does not drop. These savings come from about 40\% fewer language-model calls, since the agent now hands over to the retrained policy earlier in each task. We further conduct a second round of self-improvement on 30 randomly selected tasks with new system rollout data $\mathcal{D}_2$. As shown in \cref{fig:selfimprove}b, this round lowers the API cost and time per episode by a further 18\% to 27\%, while the success rate stays above 98\% in both settings, which suggests that self-improvement can be repeated as the system runs. Details about the cost of data collection are \mbox{provided in \appref{sec:supp_stage2}.}

\paragraph{Curated system rollouts are as effective as expert demonstrations}
We then ask whether curated system rollouts can replace expert demonstrations for training a policy to complete whole tasks. We therefore train two policies on the same amount of data from randomized scenes, one on curated system rollouts and one on official expert demonstrations, and test both without the agent in the easy and hard settings. The two policies achieve similar success rates (see \appref{sec:supp_stage2}), within 3.0 percentage points in each setting. This suggests that our system can also collect training data for visuomotor policies, as an alternative to teleoperation or \mbox{scripted motion planning.}

\subsection{Boundary analysis}
\label{sec:boundary}

\begin{figure*}[t]
\centering
\includegraphics[width=\textwidth]{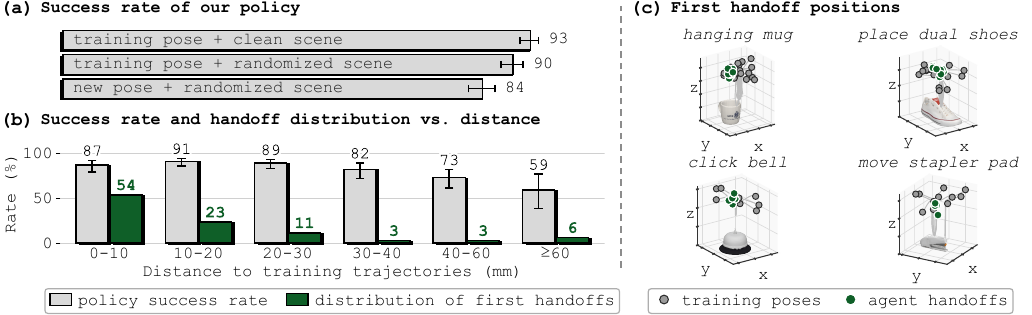}
\caption{\textbf{The agent hands over where the policy is reliable.} (a) Success rate of our policy alone on the 147 subtasks of $\mathcal{D}_0$ in three settings. (b) Success rate of the policy (gray) and distribution of the agent's first handoffs (green) over the distance to the training trajectories. (c) Training poses and first handoffs of the agent in four tasks. Error bars show 95\% Wilson intervals.}
\label{fig:boundary}
\end{figure*}

To understand where the gains of boundary-explicit training come from, we analyze the handoff boundary in our system and in Harness VLA, which leads to three key findings.

\paragraph{Key Finding 1: The policy is reliable near its training data}
To assess how our policy performs inside its handoff boundary, we evaluate it on the 147 subtasks of $\mathcal{D}_0$ after boundary-explicit training and report the success rates in \cref{fig:boundary}a. Starting from training poses, the policy reaches 93.2\% in clean scenes and 89.8\% in randomized scenes, so it generalizes well across scene appearance. Furthermore, starting from new poses, namely held-out start poses and their shifts by up to 4\,cm, its success rate stays high at 83.7\%. \appref{sec:supp_boundary} details the evaluation settings. Overall, the policy is robust to pose and scene changes near its training start poses, which serves as its handoff boundary for the agent.

\paragraph{Key Finding 2: The agent hands over where the policy is reliable}
To check where the agent hands over, we measure the distance from the gripper at each first handoff to the training trajectories of the policy, and group the trials of the policy from new poses by the same distance. As shown in \cref{fig:boundary}b, 91.1\% of the first handoffs lie within 40\,mm, where the policy succeeds in at least 82.2\% of trials. Only 6.1\% of the first handoffs lie beyond 60\,mm, where the policy still succeeds in 59.1\% of trials. \cref{fig:boundary}c further shows that the first handoffs in four tasks lie among the training trajectories. Taken together, these results show that, under the handoff protocol, the agent mostly hands over within the handoff boundary of the policy.

\paragraph{Key Finding 3: Without an explicit handoff boundary, the agent cannot hand over reliably}
We also apply the same analysis to Harness VLA, whose agent calls a whole-task policy, and find two problems. \textbf{First}, the agent calls the policy from the initial pose in almost all episodes, which leaves the whole task to the policy, and only 71.5\% of these episodes succeed. \textbf{Second}, when the agent corrects the arm and calls the policy again, the gripper is on average 80.1\,mm from the training trajectories, and only 36.8\% of such episodes succeed. As a result, the agent either leaves the whole task to the policy or hands over far from its training data, which limits the system to a success rate of 72.0\%.

\begin{figure}[t]
\centering
\includegraphics[width=\columnwidth]{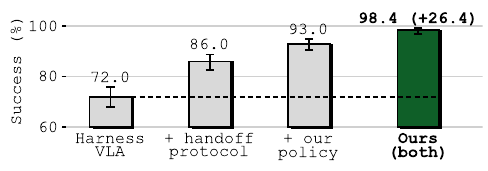}
\caption{\textbf{Ablations on the policy and the handoff protocol.} \mbox{\texttt{+ handoff protocol}} equips Harness VLA with our handoff protocol while keeping its whole-task policy, and \mbox{\texttt{+ our policy}} equips it with our policy. Success rates on all 50 tasks of RoboTwin 2.0 (C2R). Error bars show \mbox{95\% Wilson intervals.} The success axis starts at 60\%.}
\label{fig:ablation}
\vspace{2.4pt}%
\end{figure}
\begin{figure*}[t]
\centering
\includegraphics[width=0.947\textwidth]{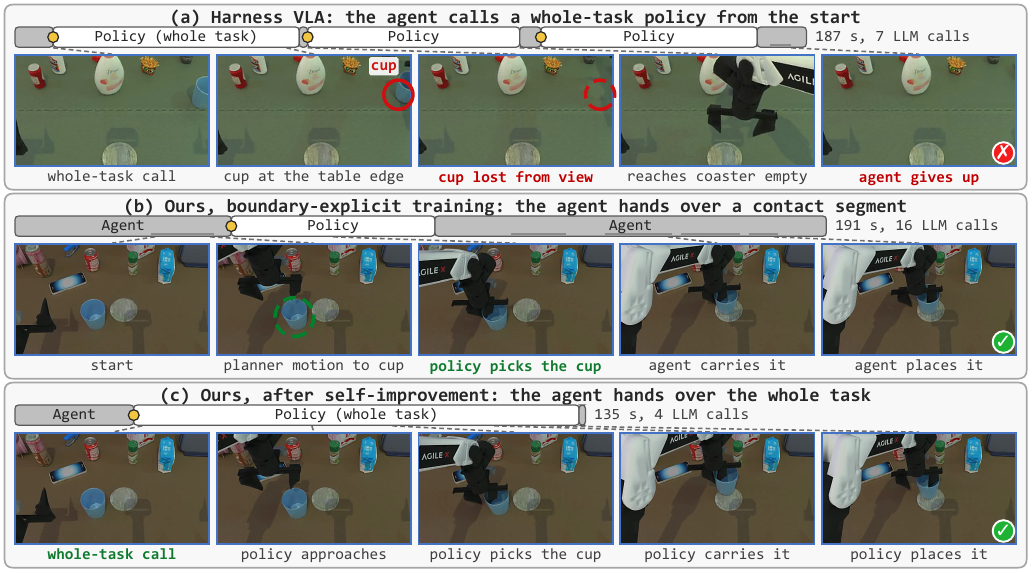}%
\caption{\textbf{Where the agent hands over decides how the system behaves.} One episode of placing a cup on a coaster for (a) Harness VLA and for our system (b) before and (c) after self-improvement, where (b) and (c) start from the same scene and (a) shows another seed. The bar above each row shows who controls the robot over time on a shared time axis: gray for the agent, with darker strips for planner motions, white for the policy, and circles for handoffs. LLM calls count the language-model calls of the episode.}
\label{fig:qual}
\end{figure*}

\subsection{Ablations}
\label{sec:ablation}

\paragraph{Both our policy and the handoff protocol contribute}
We ablate the policy and the agent in a $2\times2$ comparison. Starting from Harness VLA, we replace its whole-task policy with our policy, equip its agent with our handoff protocol, or do both, and evaluate each system on all 50 tasks of RoboTwin 2.0 under C2R, each with its own exploration. As shown in \cref{fig:ablation}, our policy alone raises the success rate of Harness VLA from 72.0\% to 93.0\%, and our handoff protocol alone raises it to 86.0\% even with the whole-task policy. With both, the system reaches 98.4\%, the highest success rate among the four systems.

\begin{table}[t]
\centering
\caption{\textbf{Our system with other language models and reasoning efforts.} Each configuration explores the tasks itself and evaluates with its own memory. We report the success rate (\%) on the 30-task subset of RoboTwin 2.0 (C2R), with 10 episodes per task, and the API cost per evaluation episode at official list prices (USD).}
\label{tab:ablation_llm}
\vspace{2pt}
\small
\setlength{\tabcolsep}{4pt}
\renewcommand{\arraystretch}{1.0}
\begin{tabularx}{\columnwidth}{@{}l *{3}{>{\centering\arraybackslash}X}@{}}
\toprule
\textbf{Language model} & \textbf{Effort} & \textbf{Success} & \textbf{Cost (\$)} \\
\midrule
Claude Opus 5.5 & high & 98.7 & 1.53 \\
Claude Opus 5.5 & low & 80.3 & 0.88 \\
Claude Opus 5.5 & max & 99.0 & 3.29 \\
\midrule
GPT-6.1 Sol & high & 85.0 & 0.26 \\
GPT-6 Astra & high & 83.3 & 1.29 \\
Kimi K3.1 Pro & high & 73.3 & 1.07 \\
DeepSeek-V4.1-Flash & high & 67.7 & 0.62 \\
\bottomrule
\end{tabularx}
\vspace{4.8pt}%
\end{table}
\paragraph{Both the language model and the reasoning effort of the agent affect system performance}
We vary the language model and the reasoning effort of the agent and evaluate each configuration on 30 randomly selected tasks under C2R, which \appref{sec:supp_results} lists. The policy and the handoff protocol stay unchanged, and each configuration explores the tasks from scratch. \textbf{First}, as shown in \cref{tab:ablation_llm}, the four other language models reach 67.7\% to 85.0\% without retraining the policy, compared with 98.7\% for Claude Opus 5.5. \textbf{Second}, a low token price does not guarantee a low API cost per episode: DeepSeek-V4.1-Flash has less than one sixth of the token price of GPT-6.1 Sol\footnote{Official API list prices in USD per million tokens: GPT-6.1 Sol charges \$2.00 for input and \$10.00 for output, and DeepSeek-V4.1-Flash charges \$0.30 and \$1.20 at its peak-hour rate.} but costs \$0.62 per episode compared with \$0.26, since it fails more often and makes many more language-model calls in failed episodes. \textbf{Third}, for Claude Opus 5.5, low effort lowers the API cost per episode from \$1.53 to \$0.88 but also lowers the success rate to 80.3\%, whereas max effort reaches 99.0\%, only 0.3 percentage points above high effort, at more than twice the cost. Considering the trade-off between success rate and API cost, we use high effort in all other experiments.

\subsection{Qualitative analysis}
\label{sec:qual}

In \cref{fig:qual}, we visualize the complete execution of placing a cup on a coaster by Harness VLA and by our system before and after self-improvement and make three observations.

\paragraph{Harness VLA tends to hand over the whole task at the start and fails after several attempts}
As shown in \cref{fig:qual}a, the agent of Harness VLA calls the whole-task policy from the initial pose, as it does in almost all episodes (Key Finding 3 in \cref{sec:boundary}). In this scene, unseen in training, the policy fails to grasp the cup and knocks it off the table, so the gripper reaches the coaster empty. With the cup off the table, two further whole-task calls also fail, and \mbox{the agent gives up.}

\paragraph{After boundary-explicit training, the agent hands over only the contact segment}
\looseness=1 As shown in \cref{fig:qual}b, the agent of our system first moves the gripper above the cup with one planner motion, so that the policy starts inside its handoff boundary. The agent then hands over only the grasp with the short command ``Pick up the cup with the left arm'', and the policy lifts the cup within two chunks. The agent carries and places the cup itself with three more planner motions and a release, and \mbox{the episode succeeds.}

\paragraph{After self-improvement, the retrained policy performs the whole task in the same steps as the system}
\looseness=1 As shown in \cref{fig:qual}c, the agent calls the retrained policy with the task instruction from the initial pose. The policy then approaches, grasps, carries and places the cup in a single call, following the same steps as the whole system in \cref{fig:qual}b. This makes the system cheaper and faster, with only 4 language-model calls and 135\,s in this episode. Failure cases of our system are shown in \mbox{\appref{sec:supp_results}.}

%% file: sections/5_conclusion.tex
\section{Conclusion}
\label{sec:conclusion}

We presented Boundary-Aware Policy Learning, which enables more reliable agent--policy coordination in agent-orchestrated robot manipulation. Boundary-Explicit Training fine-tunes the policy only on contact segments with diverse start poses, so that a short handoff protocol can describe its wide and explicit handoff boundary to the agent. Boundary-Expanding Self-Improvement then retrains the policy on the system's own curated rollouts, so that the boundary extends to earlier stages of each task. With boundary-explicit training alone, the system outperforms the strongest baseline by more than 26 percentage points in randomized scenes of RoboTwin~2.0 and on RoboDojo, and self-improvement further reduces the API cost and time per episode by 41.3\% and 37.5\% on average without loss in success rate. These results suggest that an orchestrated system improves most when the boundary between the agent and the policy is made explicit and expanded over time.

\paragraph{Limitations and future work}
\textbf{First}, data curation discards the suboptimal segments of each rollout, which DAgger~\cite{ross2011dagger} or offline reinforcement learning~\cite{levine2020offline} may turn into training data on how the system recovers from mistakes. \textbf{Second}, the handoff protocol describes the boundary only by rules, and a policy that outputs a confidence for the current state, like SayCan~\cite{saycan2022}, may let the agent check \mbox{each handoff position.}

%% file: sections/6_llm_usage_statement.tex
\section*{LLM Usage Statement}

Besides the LLMs in our system, we used LLMs to assist with coding and to edit and proofread the text for grammar, clarity and readability. All LLM-assisted code and text were carefully reviewed and revised by the authors, who made all research decisions and take full responsibility for the \mbox{content of this paper.}

%% file: sections/A_appendix.tex
\clearpage
\onecolumn
\appendix
\crefalias{section}{appendix}
\section*{Appendix}

\input{sections/supp/overview}

\input{sections/supp/sec_A_setup}

\input{sections/supp/sec_B_stage1}

\input{sections/supp/sec_C_stage2}

\FloatBarrier%
\input{sections/supp/sec_D_boundary}
\FloatBarrier%
\input{sections/supp/sec_E_results}
\FloatBarrier%
\input{sections/supp/sec_F_prompt}

\IfFileExists{sections/supp/sec_closer_look.tex}{\input{sections/supp/sec_closer_look}}{}

%% file: sections/supp/overview.tex
\noindent The appendix is organized as follows.
\begin{itemize}[leftmargin=*,itemsep=1pt,topsep=2pt]
  \item \textbf{Experimental Setup} (\cref{sec:supp_setup}): the benchmarks and the evaluation protocol, the tools, observations and memory of the agent, and the baselines.
  \item \textbf{Details of Boundary-Explicit Training} (\cref{sec:supp_stage1}): the object-interaction data $\mathcal{D}_0$ and its subtasks on both benchmarks, the training of our policy, and the handoff protocol in the prompt of the agent.
  \item \textbf{Details of Boundary-Expanding Self-Improvement} (\cref{sec:supp_stage2}): the data curation pipeline, retraining and re-exploration, the cost of data collection, the easy and hard evaluation settings, the numbers behind \mref{fig:selfimprove}, and the comparison of policies trained on curated system rollouts and on expert demonstrations.
  \item \textbf{Details of the Boundary Analysis} (\cref{sec:supp_boundary}): the evaluation of our policy on the 147 subtasks, the distance to the training trajectories, and the analysis of Harness VLA.
  \item \textbf{Additional Results} (\cref{sec:supp_results}): the task subset of the ablations, the scores on RoboDojo, the failure cases of our system, and a first attempt to deploy our system on a real robot (\cref{sec:supp_real}).
  \item \textbf{Prompt and Memory} (\cref{sec:supp_prompt}): the evaluation prompt of the agent after boundary-explicit training, shown as the sections that we add for the handoff protocol and the sections kept from Harness VLA, and two entries of its memory.
  \IfFileExists{sections/supp/sec_closer_look.tex}{\item \textbf{A Closer Look at Agent-Direct Control} (\cref{sec:supp_closer_look}): an early exploratory study of agents that control the robot directly, which reports what the agent can and cannot do and discusses future directions.}{}
\end{itemize}

%% file: sections/supp/sec_A_setup.tex
\section{Experimental Setup}
\label{sec:supp_setup}

This section details the setup of the experiments in \mref{sec:exp}: the benchmarks and the evaluation protocol, the agent with its tools, observations and memory, and the baselines.

\begin{figure}[!htb]
\centering
\includegraphics[width=\linewidth]{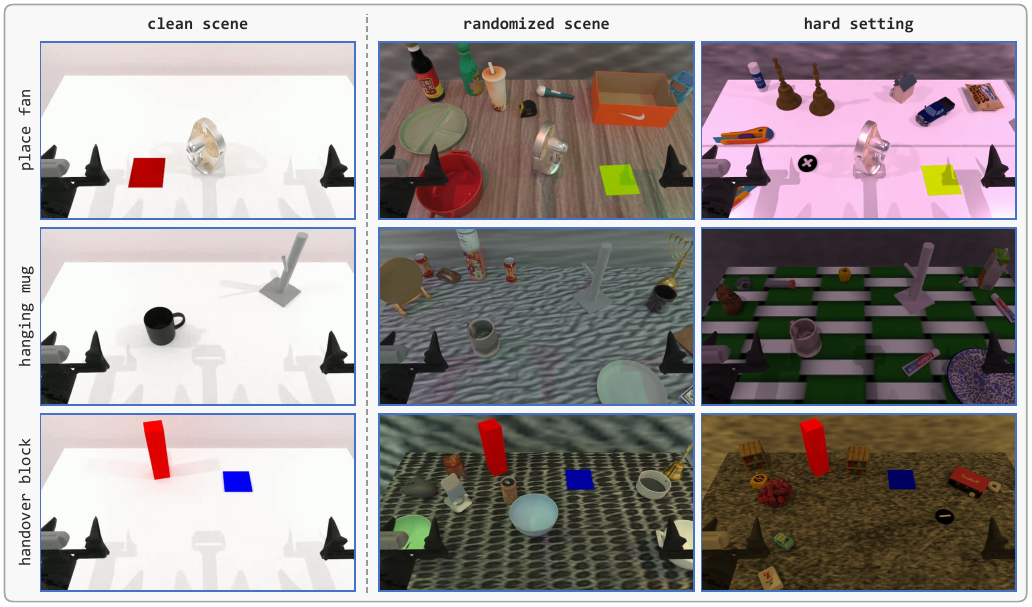}
\caption{\textbf{Clean, randomized and hard scenes.} Head-camera images at the start of an episode for three tasks. The randomized scene is built with the official randomized configuration, and the hard setting (\cref{sec:supp_stage2}) builds the same scene with textures, distractor objects and lighting that do not occur during data collection or exploration. The clean scene shows another layout of the same task.}
\label{fig:supp_settings}
\end{figure}

\paragraph{Benchmarks.}
RoboTwin~2.0~\cite{robotwin2_2025} contains 50 tasks on a dual-arm robot. Compared with its clean configuration, the official randomized configuration changes the background textures, scatters distractor objects on the table, randomizes the lighting and lowers the table by a random height, as \cref{fig:supp_settings} illustrates. On RoboDojo~\cite{chen2026robodojo}, we use the 26 tasks of the generalization, memory and open-vocabulary subsets listed in \cref{tab:supp_robodojo}, and exclude the long-horizon and precision subsets, on which training a policy rarely succeeds. Each generalization task has a clean variant and a randomized variant, which replaces the main objects with unseen instances and adds distractor objects in most tasks.

\input{sections/supp/tab_robodojo_tasks}

\paragraph{Exploration and evaluation protocol.}
On RoboTwin~2.0, exploration follows Harness VLA~\cite{harnessvla2026}: each task is explored in a clean scene in up to three agent sessions with up to five attempts each, the scene is reset before every attempt, and every new session reads the failure records of the previous sessions. The exploration of a task stops at its first success, and the notes of all tasks are then merged into one memory, which stays frozen during evaluation. Our system and every baseline are evaluated on 10 different scene seeds per task, each baseline in its own configuration. In the main results, an episode of our system is limited to 100 language-model calls and 5{,}000\,s, and the native predicate of the benchmark judges success. The easy and hard settings of self-improvement are described in \cref{sec:supp_stage2}.

\paragraph{Agent and tools.}
The agent is Claude Opus 5.5 with high reasoning effort and acts only through registered tools, the main ones of which are listed in \cref{tab:supp_tools}, since we disable the built-in tools of the agent framework, such as the shell. Perception tools return camera images, the image location of a named object from SAM~3~\cite{carion2025sam3} and the world coordinates of image pixels, and motion tools move one end effector to a pose through the benchmark's motion planner, rotate a wrist or set a gripper. The policy tool takes a command $u$ and a number of chunks, which the tool limits to at most four after boundary-explicit training and to at most 24 after self-improvement (\cref{sec:supp_stage2}). Each chunk consists of 50 actions, which specify the absolute end-effector poses and gripper commands of both arms and are executed for 40\,ms each. The policy observes the scene again before every chunk, and the call stops as soon as the native predicate reports success. The reset tool exists only during exploration.

\input{sections/supp/tab_tools}

\paragraph{Observations and success signal.}
The agent and the policy observe the scene through the same three RGB cameras, a head camera and two wrist cameras, which also record $\mathcal{D}_0$. These cameras render $640\times480$ images with wider vertical fields of view than the benchmark's default cameras, namely 55\textdegree{} at the head and 85\textdegree{} at the wrists. After every tool call that changes the robot state, the agent receives the three images together with the proprioception, which reports the joint positions, the end-effector poses of both arms and the gripper openings. Depth is never shown to the agent as an image, and the agent instead reads the world coordinates of selected pixels or of an image region. Besides these observations, the agent receives the binary success flag of the benchmark's native predicate, which an operator or a success detector would provide on a real robot. The tool that ends an episode reports the result of this predicate, so the agent cannot declare success by itself. Simulator state such as object poses or contacts reaches neither the agent nor the policy, and the policy does not see the success flag either.

\paragraph{Memory.}
The memory has three layers: global notes that apply to all tasks, task-family notes, and for every task a record of its successful exploration attempt. This record contains the plan of the attempt and its recipe, the sequence of state-changing commands with their arguments, with an example in \cref{sec:supp_prompt}. During exploration, the prompt adds the memory-writing instructions of Harness VLA: before every reset, the agent writes a failure record that names the one change it will make in the next attempt, and after a success, it writes the record of the final attempt. At evaluation, the frozen memory is read-only, and the prompt asks the agent to use the recipe as a plan for the order and type of actions, not as coordinates to replay.

\paragraph{Baselines.}
On RoboTwin~2.0, the whole-task policy $\pi_{0.5}$~\cite{pi05_2025} uses the model and optimization settings of our policy (\cref{sec:supp_stage1}), is fine-tuned on $\mathcal{D}_0$ together with 2{,}650 official whole-task demonstrations in clean scenes (53 per task), and runs end to end from the task instruction. On RoboDojo, $\pi_{0.5}$ is fine-tuned on the official demonstrations, which contain none of the open-vocabulary tasks. AGP~\cite{jia2026agp} observes the scene only through RGB images and writes programs for perception and robot control at run time, while ORA~\cite{ora2026}, which we run with its official implementation, gives the agent predefined perception and motion tools. As summarized in \cref{tab:supp_baselines}, neither AGP nor ORA uses a learned policy, explores the tasks or keeps a memory. For Harness VLA~\cite{harnessvla2026} on RoboTwin~2.0, we run its official implementation with the LingBot-VLA~\cite{lingbotvla2026} checkpoint released with it and with a memory from its own exploration, under the same per-episode limits as our system. On RoboDojo, we follow its official setting and equip it with VPP2~\cite{guo2026vpp2}. All agent-based methods that we run use Claude Opus 5.5 with high reasoning effort.

\input{sections/supp/tab_baselines}

\paragraph{ORA on RoboDojo.}
ORA is not evaluated on RoboDojo in \mref{tab:main}, since its agent often specifies targets that the robot cannot reach. We test the early version of ORA that, according to its public description, was used in preliminary runs on RoboDojo. Its motion back end mainly solves inverse kinematics for a limited set of candidate end poses and checks joint-interpolated paths, and it does not plan detour trajectories. After a failed motion, the agent therefore has to choose another pose or split the motion into smaller steps, whereas our system and Harness VLA plan complete trajectories with a motion planner. In the episodes that we analyze from several tasks, a large share of the motions of ORA do not complete because of failed inverse kinematics, joint jumps, self-collision checks or targets that are not reached, which repeatedly interrupts multi-stage manipulation.

\paragraph{Comparison with related systems.}
In \cref{tab:supp_related}, we compare our system with three systems in which an agent orchestrates a learned policy. Harness VLA and BATON~\cite{xu2026baton} keep the policy frozen and let the agent figure out when and where to hand over through exploration and record what works in a memory, while Robo-COP~\cite{robocop2026} fine-tunes the policy on its own executions to fix recurring failures. In boundary-explicit training, we fine-tune the policy only on contact segments with diverse start poses around the target object, so that a short handoff protocol describes its handoff boundary to the agent. Among these systems, we use Harness VLA as the baseline, since it releases its implementation together with a memory and a policy for RoboTwin~2.0.

\input{sections/supp/tab_related}

%% file: sections/supp/tab_robodojo_tasks.tex
\begin{table}[!htb]
\centering
\caption{\textbf{The 26 RoboDojo tasks used in our evaluation.} Each generalization task is evaluated in its randomized variant.}
\label{tab:supp_robodojo}
\vspace{2pt}
\small
\setlength{\tabcolsep}{5pt}
\renewcommand{\arraystretch}{1.15}
\begin{tabular}{@{}l c >{\raggedright\arraybackslash}p{0.66\linewidth}@{}}
\toprule
\textbf{Subset} & \textbf{Tasks} & \textbf{Task names} \\
\midrule
Generalization & 12 & \mbox{\textit{stack bowls}}, \mbox{\textit{push T}}, \mbox{\textit{pack objects into box}}, \mbox{\textit{fold clothes}}, \mbox{\textit{hang mugs}}, \mbox{\textit{sweep blocks}}, \mbox{\textit{pour liquid into cup}}, \mbox{\textit{make toast}}, \mbox{\textit{arrange largest number}}, \mbox{\textit{sort nesting dolls by size}}, \mbox{\textit{store laptop and headphones}}, \mbox{\textit{stack blocks}} \\
\addlinespace[2pt]
Memory & 6 & \mbox{\textit{cover blocks}}, \mbox{\textit{match and pick from conveyor}}, \mbox{\textit{swap blocks}}, \mbox{\textit{swap T}}, \mbox{\textit{press by number}}, \mbox{\textit{imitate sorting sequence}} \\
\addlinespace[2pt]
Open-vocabulary & 8 & \mbox{\textit{align blocks}}, \mbox{\textit{general pickup}}, \mbox{\textit{solve equation}}, \mbox{\textit{stack blocks by language}}, \mbox{\textit{classify objects by language}}, \mbox{\textit{pick from conveyor by image}}, \mbox{\textit{store tools in toolbox}}, \mbox{\textit{pour by language}} \\
\bottomrule
\end{tabular}
\end{table}

%% file: sections/supp/tab_tools.tex
\begin{table}[!htb]
\centering
\caption{\textbf{Main tools available to the agent after boundary-explicit training.} Every tool that changes the robot state returns a fresh observation with the robot state and the images of the head camera and the two wrist cameras. Depth is never returned as an image, and the agent reads the world coordinates of chosen pixels through the two depth-query tools.}
\label{tab:supp_tools}
\vspace{2pt}
\small
\setlength{\tabcolsep}{4pt}
\renewcommand{\arraystretch}{1.12}
\begin{tabularx}{\linewidth}{@{}l l >{\raggedright\arraybackslash}X >{\raggedright\arraybackslash}X@{}}
\toprule
\textbf{Tool} & \textbf{Category} & \textbf{Arguments} & \textbf{Returns} \\
\midrule
\texttt{view\_env\_state} & Perception & step & robot state, episode status, task instruction, head and wrist RGB \\
\texttt{render} & Perception & none & a fresh synchronized observation \\
\texttt{segment} & Perception & text prompt, view & score, pixel box and mask center of the object (SAM~3~\cite{carion2025sam3}) \\
\texttt{sample\_world\_xyz} & Depth query & view, pixels & world $xyz$ (m) at each pixel \\
\texttt{query\_world\_map} & Depth query & view, pixel box & sampled world $xyz$ (m) inside the box \\
\midrule
\texttt{move\_to} & Motion & arm, end-effector position, orientation (optional), gripper & reached pose, distance to the target, plan status \\
\texttt{rotate\_wrist} & Motion & arm, yaw change & as \texttt{move\_to} \\
\texttt{set\_gripper} & Gripper & arm, opening & execution receipt \\
\texttt{release} & Gripper & arm & execution receipt \\
\midrule
\texttt{pi05\_act} & Policy & command $u$, number of chunks & executed actions, stop reason, success flag \\
\midrule
\texttt{read\_text\_file} & Memory & path & file content \\
\texttt{write\_text\_file} & Memory & path, content & none \\
\texttt{list\_dir} & Memory & path & file names \\
\midrule
\texttt{finish} & Episode & status, summary & result of the native predicate \\
\texttt{reset} & Episode & none & a new episode (exploration only) \\
\bottomrule
\end{tabularx}
\end{table}

%% file: sections/supp/tab_baselines.tex
\begin{table}[!htb]
\centering
\caption{\textbf{Configuration of the compared methods on RoboTwin~2.0.} Learned policy is the policy that the method runs, and agent control is how the agent acts on the robot.}
\label{tab:supp_baselines}
\vspace{2pt}
\small
\setlength{\tabcolsep}{8pt}
\renewcommand{\arraystretch}{1.12}
\begin{tabular}{@{}llll@{}}
\toprule
\textbf{Method} & \textbf{Learned policy} & \textbf{Agent control} & \textbf{Exploration and memory} \\
\midrule
$\pi_{0.5}$~\cite{pi05_2025} & whole-task $\pi_{0.5}$ & none & none \\
AGP~\cite{jia2026agp} & none & runtime programming & none \\
ORA~\cite{ora2026} & none & predefined tools & none \\
Harness VLA~\cite{harnessvla2026} & whole-task LingBot-VLA & predefined tools & own exploration, frozen memory \\
\rowcolor{oursrow}
\oursL{\textbf{Ours}} & our policy ($\pi_{0.5}$ on $\mathcal{D}_0$) & predefined tools & \multicolumn{1}{>{\columncolor{oursrow}[\tabcolsep][0pt]}l@{}}{own exploration, frozen memory} \\
\bottomrule
\end{tabular}
\end{table}

%% file: sections/supp/tab_related.tex
\begin{table}[!htb]
\centering
\caption{\textbf{Comparison with related systems in which an agent calls a learned policy.} The third column states how the agent learns where to hand over, and the last column states what each system updates from its own executions.}
\label{tab:supp_related}
\vspace{2pt}
\small
\setlength{\tabcolsep}{8pt}
\renewcommand{\arraystretch}{1.12}
\begin{tabular}{@{}llll@{}}
\toprule
\textbf{System} & \textbf{Policy} & \textbf{Where to hand over is learned from} & \textbf{Updated from own executions} \\
\midrule
Harness VLA~\cite{harnessvla2026} & frozen & exploration & memory \\
BATON~\cite{xu2026baton} & frozen & subtask exploration & memory \\
Robo-COP~\cite{robocop2026} & fine-tuned on own executions & deployment & policy and memory \\
\rowcolor{oursrow}
\oursL{\textbf{Ours}} & fine-tuned on contact segments & handoff protocol and exploration & \multicolumn{1}{>{\columncolor{oursrow}[\tabcolsep][0pt]}l@{}}{policy and memory} \\
\bottomrule
\end{tabular}
\end{table}

%% file: sections/supp/sec_B_stage1.tex
\section{Details of Boundary-Explicit Training}
\label{sec:supp_stage1}

This section details the object-interaction data $\mathcal{D}_0$, the training of our policy and the handoff protocol of boundary-explicit training in \mref{sec:aligned_learning}.

\paragraph{Object-interaction data.}
We generate $\mathcal{D}_0$ in clean scenes with the scripted expert of RoboTwin~2.0, which uses the native motion planner and grasp poses of the benchmark, and keep only successful demonstrations.
Each demonstration belongs to one of 147 subtasks, which are defined by the task, the object, the role of the object in the task and the skill, and these subtasks are summarized by their 24 skills in \cref{tab:supp_verbs}.
Recording starts at a sampled start pose of the gripper near the object, after an unrecorded scripted approach, and ends once a picked object has been lifted by 7\,cm and held for 0.4\,s or once the native motion of any other skill ends.
For grasping skills, the start pose places the gripper 8 to 12\,cm above a feasible grasp point, either 4 to 8\,cm to one side of it or within a 2\,cm disk above it, and rotates the gripper about the vertical axis by $\pm5^\circ$ or $\pm15^\circ$ from the native grasp orientation.
The handoff protocol in \mref{sec:protocol} restates the height and the yaw variation of this rule to the agent.

\input{sections/supp/tab_d0_verbs}

\paragraph{Dataset and policy training.}
$\mathcal{D}_0$ contains 2{,}699 training demonstrations, or 2.9 hours at 25\,Hz, and one further demonstration of each subtask, 147 in total, is held out for validation.
Each demonstration pairs one command $u$ that names the skill, the object and the arm, such as ``Turn the bottle upright with the left arm'', with the images of the three cameras, the proprioceptive state and the absolute end-effector actions at every step.
We fine-tune all parameters of $\pi_{0.5}$ from its base checkpoint with its original training objective, and change only the data and the action representation.
The state and the action of each arm consist of a tool center point (TCP) position and a 6D rotation in the world frame, together with the gripper opening, and the agent specifies its motion targets in the same frame.
We train with the hyperparameters in \cref{tab:supp_hparams} and use the checkpoint with the lowest loss on the validation demonstrations, so the evaluation episodes play no role in this choice.

\input{sections/supp/tab_hparams}

\paragraph{Object-interaction data on RoboDojo.}
RoboDojo does not release a scripted expert, so we write one for its atomic skills and generate $\mathcal{D}_0$ by the same procedure as on RoboTwin~2.0.
The approach to a start pose near the object is not recorded, and recording covers only the contact segment that follows.
This $\mathcal{D}_0$ contains 3{,}464 training and 182 validation demonstrations of 112 subtasks, which \cref{tab:supp_verbs_robodojo} summarizes by their 11 skills.
The training configuration is listed in \cref{tab:supp_hparams} and differs from that on RoboTwin~2.0 in the number of training steps and in the actions, which are expressed relative to the end-effector pose at the start of each chunk.

\input{sections/supp/tab_d0_robodojo}

\paragraph{Handoff protocol in the prompt.}
The rules of the handoff protocol in \mref{sec:protocol} are written in fixed sections of the prompt (\cref{sec:supp_prompt}), and the text of these sections is the same for every task.
The section on the local-skill handoff tells the agent to hand over grasping and local object manipulation to the policy with one short command, and to use the motion tools for approach, transport and recovery.
The section on pose suggestions describes how to prepare a call, with a grasp-compatible orientation and the fingertip center roughly 8 to 12\,cm above the grasp region, and presents these values as defaults derived from the training data. When the gripper is already at such a pose, the agent may call the policy directly.
The same section suggests three or four chunks for a clear pickup and one or two near precise contact, and it asks the agent to treat an object as held only when the object moves with the gripper.
The policy tool enforces the limit of four chunks per call (\cref{sec:supp_setup}), and exploration and evaluation use the same sections of the handoff protocol.

%% file: sections/supp/tab_d0_verbs.tex
\begin{table}[!htb]
\centering
\caption{\textbf{The 147 subtasks of the object-interaction data $\mathcal{D}_0$, summarized by skill.} Each row gives the number of subtasks and of training demonstrations of one skill.}
\label{tab:supp_verbs}
\vspace{2pt}
\small
\setlength{\tabcolsep}{5pt}
\renewcommand{\arraystretch}{1.12}
\begin{tabular}{@{}lcc@{\hspace{28pt}}lcc@{}}
\toprule
\textbf{Skill} & \textbf{Subtasks} & \textbf{Demonstrations} & \textbf{Skill} & \textbf{Subtasks} & \textbf{Demonstrations} \\
\cmidrule(r){1-3}\cmidrule(l){4-6}
pick lift & 104 & 1{,}957 & place in held skillet & 1 & 19 \\
insert into open drawer & 7 & 86 & pull drawer & 1 & 19 \\
handover & 5 & 71 & rotate place sign & 1 & 19 \\
dual pick lift & 4 & 76 & toggle switch & 1 & 19 \\
press retract & 3 & 57 & pour into bigbin & 1 & 19 \\
stack on block & 3 & 57 & hang handle & 1 & 19 \\
nest on bowl & 3 & 56 & open door & 1 & 19 \\
pick lift regrasp & 2 & 35 & seat phone in stand & 1 & 19 \\
orient bottle & 1 & 19 & align scan & 1 & 19 \\
strike block & 1 & 19 & shake vertical & 1 & 19 \\
move away local & 1 & 19 & shake horizontal & 1 & 19 \\
open lid & 1 & 19 & stamp contact place & 1 & 19 \\
\midrule
\multicolumn{4}{@{}l}{\textbf{Total (24 skills)}} & \textbf{147} & \textbf{2{,}699} \\
\bottomrule
\end{tabular}
\end{table}

%% file: sections/supp/tab_hparams.tex
\begin{table}[!htb]
\centering
\caption{\textbf{Training configuration of our policy on RoboTwin~2.0 and RoboDojo.} A value that spans both columns holds for both benchmarks. The training time is measured up to the checkpoint that we use.}
\label{tab:supp_hparams}
\vspace{2pt}
\small
\setlength{\tabcolsep}{6pt}
\renewcommand{\arraystretch}{1.12}
\begin{tabular}{@{}lll@{}}
\toprule
\textbf{Setting} & \textbf{RoboTwin~2.0} & \textbf{RoboDojo} \\
\midrule
\multicolumn{3}{@{}l}{\textit{Data}} \\
Training demonstrations & 2{,}699 contact segments & 3{,}464 contact segments \\
Scenes & \multicolumn{2}{l}{clean} \\
Action interval & \multicolumn{2}{l}{40\,ms (25\,Hz)} \\
\midrule
\multicolumn{3}{@{}l}{\textit{Model}} \\
Base model & \multicolumn{2}{l}{$\pi_{0.5}$ base checkpoint} \\
Trainable parameters & \multicolumn{2}{l}{all (3.35\,B)} \\
Cameras & \multicolumn{2}{l}{head and two wrist RGB cameras, $224\times224$} \\
State & \multicolumn{2}{l}{per arm, absolute world-frame end-effector position, 6D rotation and gripper opening} \\
Action & absolute, in the format of the state & relative to the pose at the start of the chunk \\
Action chunk & \multicolumn{2}{l}{50 actions} \\
\midrule
\multicolumn{3}{@{}l}{\textit{Optimization}} \\
Global batch size & \multicolumn{2}{l}{1{,}024} \\
Learning rate & \multicolumn{2}{l}{peak $1\times10^{-4}$, cosine decay} \\
Optimizer & \multicolumn{2}{l}{AdamW ($\beta_1{=}0.9$, $\beta_2{=}0.95$), gradient clipping at 1.0} \\
EMA decay & \multicolumn{2}{l}{0.99} \\
Training steps & 800 & 1{,}800 \\
Hardware & \multicolumn{2}{l}{16 GPUs} \\
Training time & 1.7\,h & 3.8\,h \\
\bottomrule
\end{tabular}
\end{table}

%% file: sections/supp/tab_d0_robodojo.tex
\begin{table}[!htb]
\centering
\caption{\textbf{The 112 subtasks of $\mathcal{D}_0$ on RoboDojo, summarized by skill.} Each row gives the number of subtasks and of training demonstrations of one skill.}
\label{tab:supp_verbs_robodojo}
\vspace{2pt}
\small
\setlength{\tabcolsep}{5pt}
\renewcommand{\arraystretch}{1.12}
\begin{tabular}{@{}lcc@{\hspace{28pt}}lcc@{}}
\toprule
\textbf{Skill} & \textbf{Subtasks} & \textbf{Demonstrations} & \textbf{Skill} & \textbf{Subtasks} & \textbf{Demonstrations} \\
\cmidrule(r){1-3}\cmidrule(l){4-6}
pick up & 99 & 3{,}089 & lower bread into toaster slot & 1 & 29 \\
press button & 4 & 124 & close lid & 1 & 29 \\
insert coin & 1 & 29 & screw nut onto bolt & 1 & 29 \\
insert charger into socket & 1 & 29 & rotate and hand over & 1 & 29 \\
insert test tube into rack & 1 & 29 & put nut onto bolt & 1 & 19 \\
lower ring onto pole & 1 & 29 & & & \\
\midrule
\multicolumn{4}{@{}l}{\textbf{Total (11 skills)}} & \textbf{112} & \textbf{3{,}464} \\
\bottomrule
\end{tabular}
\end{table}

%% file: sections/supp/sec_C_stage2.tex
\section{Details of Boundary-Expanding Self-Improvement}
\label{sec:supp_stage2}

This section gives the details of boundary-expanding self-improvement, which \mref{sec:expand} describes and \mref{sec:expandexp} evaluates.

\paragraph{Data curation.}
We collect the system rollouts on RoboTwin~2.0 with the system after boundary-explicit training under the official randomized configuration and keep only the successful rollouts.
A curation agent, which is a language model, then reads each rollout through its tool calls with their arguments and results, the gripper state from proprioception and head-camera images taken at key moments of the rollout, such as after each grasp and release. Apart from the success flag, no simulator state such as object poses reaches the curation agent. For every grasp in the rollout, the curation agent decides whether the grasp is part of the solution that finally succeeds or belongs to a failed or repeated attempt, and marks the calls that only adjust the arm pose. The suboptimal segments are these failed and repeated attempts and pose adjustments. These segments are removed, and the remaining calls form the reference in their original order. Fixed rules then merge repeated calls, may combine an earlier grasp with the transport and placement of the final attempt, and restore or add a free move where the removal leaves a large gap between kept poses.

\paragraph{Re-execution and acceptance.}
A rollout without suboptimal segments is kept as it is, and every other rollout is re-executed. For re-execution, the scene is reset to its initial state, and the agent $\mathcal{A}$ follows the reference step by step while the policy is called online, with up to three attempts. A re-executed rollout enters $\mathcal{D}_1$ only if the task succeeds, the executed steps follow the order of the reference and no grasp is empty. Rollouts that are unusually long, or in which an object that the robot did not touch moves, are also discarded. The latter check reads object poses after collection and only filters the data: no simulator state reaches the agent, the curation agent or the policy. After curation, $\mathcal{D}_1$ contains 30 rollouts per task, or 1{,}500 in total.

\paragraph{Retraining and re-exploration.}
We retrain the policy from the base checkpoint of $\pi_{0.5}$ on $\mathcal{D}_0$ and $\mathcal{D}_1$ for 2{,}000 steps with the other settings of \cref{tab:supp_hparams}. \mbox{Besides} the whole-task rollouts, the contact segments of the system rollouts also serve as training samples with their commands $u$. After retraining, the handoff protocol is updated to describe the expanded boundary. The agent first calls the policy from the initial pose with the task instruction, and one call may now run up to 24 chunks. If the task is unfinished after this call, the agent calls the policy again with the task instruction while the task progresses. When the progress stalls or a step fails, the agent continues as in the first stage and hands over the remaining contact segments with their commands. With this protocol, the agent re-explores each task in randomized scenes from the memory of the first stage and revises this memory, which is then frozen for evaluation.
The second round on the 30-task subset (\cref{sec:supp_results}) repeats these steps: we collect $\mathcal{D}_2$ in the same way, with 30 rollouts per task, retrain the policy on $\mathcal{D}_0$, $\mathcal{D}_1$ and $\mathcal{D}_2$, and let the agent re-explore each task.

\paragraph{Cost of data collection.}
To reduce the cost of collecting $\mathcal{D}_1$ and $\mathcal{D}_2$, the agent runs on GPT-6.1 Terra, a cheaper language model, for the easy tasks and stays on Claude Opus 5.5 for the hard ones. GPT-6.1 Terra collects about 94\% of the rollouts, and the collection of both datasets costs about \$1{,}013 in API usage in total.

\paragraph{Easy and hard settings.}
Both settings evaluate each task over 10 episodes, on scene seeds that appear neither in the collection of $\mathcal{D}_1$ nor in exploration. The easy setting keeps the official randomized configuration. The hard setting replaces the wall and table textures and the distractor objects with assets imported from RoboDojo, namely 240 textures and 244 objects from 174 categories, as shown in the rightmost column of \cref{fig:supp_settings}. The hard setting also lights each scene with one of 24 lighting presets, 8 color tones at 3 intensities, which differ from all lighting seen during collection and exploration.

\paragraph{Results of self-improvement.}
We list the values behind \mref{fig:selfimprove} in \cref{tab:supp_selfimprove}, where the language-model calls, API cost and time are means per evaluation episode. The reductions of 41.3\% in API cost and 37.5\% in time per episode in \mref{sec:expandexp} are the means of the reductions in the easy and hard settings on all 50 tasks.

\input{sections/supp/tab_selfimprove}

\paragraph{Policy without the agent.}
In the comparison of \mref{sec:expandexp}, two $\pi_{0.5}$ policies run end to end from the task instruction without the agent, with 10 episodes per task in each setting. One policy is fine-tuned on official whole-task demonstrations in randomized scenes and the other on the same number of curated system rollouts per task, as listed in \cref{tab:supp_policy_alone}. In the first round, the policy trained on curated system rollouts is an additional policy trained on $\mathcal{D}_1$ alone, not the retrained policy that the agent calls.

\input{sections/supp/tab_policy_alone}

%% file: sections/supp/tab_selfimprove.tex
\begin{table}[!htb]
\centering
\caption{\textbf{Our system before and after self-improvement on RoboTwin~2.0.} Success rate, language-model calls, API cost and time per episode in the easy and hard settings, which are the values plotted in \mref{fig:selfimprove}. The values are rounded means, and Reduction gives the per-setting reductions that \mref{fig:selfimprove} shows as integers. Each setting has 10 episodes per task.}
\label{tab:supp_selfimprove}
\vspace{2pt}
\small
\setlength{\tabcolsep}{9pt}
\renewcommand{\arraystretch}{1.08}
\newcommand{\rc}[1]{\textcolor{gray}{#1}}
\begin{tabular}{@{}llcccc@{}}
\toprule
\textbf{Setting} & \textbf{Stage} & \textbf{Success (\%)} & \textbf{LLM calls} & \textbf{Cost (\$)} & \textbf{Time (min)} \\
\midrule
\multicolumn{6}{@{}l}{\textit{All 50 tasks}} \\
\multirow{3}{*}{Easy} & Before & 96.4 & 26.3 & 1.74 & 4.21 \\
                      & After  & 97.2 & 15.3 & 0.96 & 2.44 \\
                      & \rc{Reduction (\%)} & & \rc{41.8} & \rc{44.8} & \rc{42.0} \\
\addlinespace[2pt]
\multirow{3}{*}{Hard} & Before & 93.6 & 26.2 & 1.67 & 3.95 \\
                      & After  & 94.2 & 15.7 & 1.04 & 2.65 \\
                      & \rc{Reduction (\%)} & & \rc{40.1} & \rc{37.7} & \rc{32.9} \\
\midrule
\multicolumn{6}{@{}l}{\textit{30-task subset}} \\
\multirow{3}{*}{Easy} & Before       & 97.3 & 20.7 & 1.32 & 3.13 \\
                      & After        & 98.7 & 11.0 & 0.68 & 1.57 \\
                      & Second round & 98.3 & 10.0 & 0.55 & 1.28 \\
\addlinespace[2pt]
\multirow{3}{*}{Hard} & Before       & 98.3 & 19.1 & 1.12 & 2.56 \\
                      & After        & 98.7 & 12.0 & 0.75 & 1.71 \\
                      & Second round & 98.3 &  9.9 & 0.54 & 1.30 \\
\bottomrule
\end{tabular}
\end{table}

%% file: sections/supp/tab_policy_alone.tex
\begin{table}[!htb]
\centering
\caption{\textbf{Curated system rollouts are as effective as expert demonstrations.} Success rate (\%) of two $\pi_{0.5}$ policies that run end to end without the agent on RoboTwin~2.0 in the easy and hard settings, with 10 episodes per task. Within each block, the two policies are trained on the same number of whole-task episodes per task.}
\label{tab:supp_policy_alone}
\vspace{2pt}
\small
\setlength{\tabcolsep}{9pt}
\renewcommand{\arraystretch}{1.08}
\begin{tabular}{@{}lccc@{}}
\toprule
\textbf{Whole-task training data} & \textbf{Episodes per task} & \textbf{Easy} & \textbf{Hard} \\
\midrule
\multicolumn{4}{@{}l}{\textit{All 50 tasks, first round of self-improvement}} \\
Expert demonstrations & 30 & 55.8 & 55.6 \\
\rowcolor{oursrow}
\oursL{Curated system rollouts ($\mathcal{D}_1$)} & 30 & 55.4 & \oursR{54.2} \\
\midrule
\multicolumn{4}{@{}l}{\textit{30-task subset, second round of self-improvement}} \\
Expert demonstrations & 60 & 66.7 & 67.3 \\
\rowcolor{oursrow}
\oursL{Curated system rollouts ($\mathcal{D}_1$ and $\mathcal{D}_2$)} & 60 & 69.7 & \oursR{64.7} \\
\bottomrule
\end{tabular}
\end{table}

%% file: sections/supp/sec_D_boundary.tex
\section{Details of the Boundary Analysis}
\label{sec:supp_boundary}

This section details the evaluation settings and the measurements behind the boundary analysis in \mref{sec:boundary}.

\paragraph{Policy evaluation on the 147 subtasks.}
We run our policy alone, without the agent, on the 147 subtasks of $\mathcal{D}_0$ from five start poses per subtask in each of three settings, which gives 735 trials per setting. The first two settings start from training poses, the start poses of training demonstrations, in clean scenes and in randomized scenes. The third setting starts from new poses in randomized scenes: the start pose of the validation demonstration of each subtask (\cref{sec:supp_stage1}), whose scene differs from those of the training demonstrations, and four shifts of this pose by 1 to 4\,cm, one in each horizontal direction. Success in each trial is judged by a predicate of the subtask that reads the simulator state. This predicate is used only in this evaluation, and neither the agent nor the policy has access to it.

\paragraph{Distance to the training trajectories.}
The distance in Key Finding~2 is measured in the frame of the target object, whose pose comes from the simulator and serves only for this offline analysis. In this frame, the distance of a gripper position to the training trajectories is its nearest distance to the gripper positions of the training demonstrations of the same subtask and arm before contact, where contact begins when the gripper starts to close or, for skills without a grasp, at the closest approach to the object. This distance covers only the position of the gripper, not its orientation. In \mref{fig:boundary}b and c, we use the first policy calls in the C2R evaluation episodes of our system in \mref{tab:main}. The success rate of the policy in \mref{fig:boundary}b is computed over the trials from new poses above, with the distance taken at the start pose, and covers only the subtasks that start before a grasp.

\paragraph{Training trajectories of LingBot-VLA.}
The analysis of Key Finding~3 needs the training trajectories of LingBot-VLA, whose training data are not documented. We therefore use the official whole-task demonstrations of RoboTwin~2.0 as its training trajectories, since their frame count and pose statistics closely match the normalization statistics released with the policy, and cut each demonstration before the gripper starts to close.

\begin{figure}[!htb]
\centering
\includegraphics[width=0.47\linewidth]{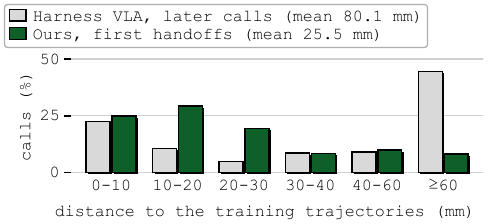}
\caption{\textbf{Later policy calls of Harness VLA start farther from the training trajectories on average than our first handoffs.} Distance to the training trajectories in the world frame for 187 calls of Harness VLA made after an arm or gripper command of the agent (gray) and the 395 first handoffs of our system (green). The legend gives the mean distances.}
\label{fig:supp_wholetask}
\end{figure}

\paragraph{Policy calls of Harness VLA.}
In the analyzed Harness VLA episodes with a policy call, the agent always makes its first call from the initial pose, and 71.5\% of these episodes succeed. In 187 later calls that the agent makes after an arm or gripper command since the previous policy call, the gripper is on average 80.1\,mm from the training trajectories, compared with 25.5\,mm for the 395 first handoffs of our system measured in the same way against $\mathcal{D}_0$ (\cref{fig:supp_wholetask}). The episodes with such a call succeed in 36.8\% of cases, compared with 90.6\% for the other episodes. Since the agent issues these commands only after an earlier policy call has left the task unfinished, these episodes may also be harder.

%% file: sections/supp/sec_E_results.tex
\section{Additional Results}
\label{sec:supp_results}

This section first gives additional results in simulation (\cref{sec:supp_results_sim}): the task subset used in \mref{sec:expandexp,sec:ablation}, the scores on RoboDojo and failure cases of our system. It then presents a first attempt to deploy our system on a real robot (\cref{sec:supp_real}).

\subsection{Additional Results in Simulation}
\label{sec:supp_results_sim}

\paragraph{Task subset of the ablations.}
We draw 30 of the 50 tasks of RoboTwin~2.0 at random: \textit{adjust bottle}, \textit{beat block hammer}, \textit{blocks ranking rgb}, \textit{click alarmclock}, \textit{click bell}, \textit{dump bin bigbin}, \textit{grab roller}, \textit{handover block}, \textit{handover mic}, \textit{lift pot}, \textit{move playingcard away}, \textit{open laptop}, \textit{pick dual bottles}, \textit{place a2b left}, \textit{place a2b right}, \textit{place bread basket}, \textit{place burger fries}, \textit{place container plate}, \textit{place empty cup}, \textit{place mouse pad}, \textit{place object scale}, \textit{place object stand}, \textit{place phone stand}, \textit{place shoe}, \textit{press stapler}, \textit{rotate qrcode}, \textit{shake bottle}, \textit{shake bottle horizontally}, \textit{stack bowls two} and \textit{stamp seal}. These 30 tasks are used in the second round of self-improvement in \mref{fig:selfimprove}b and in the language-model and reasoning-effort ablations in \mref{tab:ablation_llm}, with 10 episodes per task. Across the rows of \mref{tab:ablation_llm}, the policy and the handoff protocol stay the same, and we do not tune the prompt to any language model. The rows therefore differ only on the agent side: in the language model, in its reasoning effort, and in the memory that each configuration builds through its own exploration.

\paragraph{Scores on RoboDojo.}
Besides the success rate, RoboDojo~\cite{chen2026robodojo} defines a score for each task, and we compute it with the official implementation of the benchmark. \cref{tab:supp_robodojo_scores} lists both metrics for the three subsets. Our system also achieves the highest score in every subset.

\input{sections/supp/tab_robodojo_scores}

\begin{figure}[!htb]
\centering
\includegraphics[width=\linewidth]{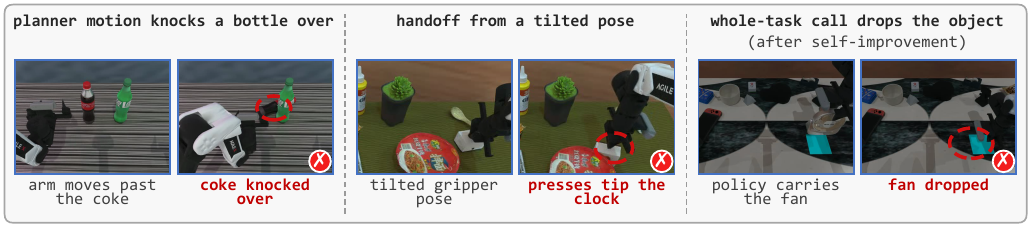}
\caption{\textbf{Failure cases of our system.} (Left) A planner motion knocks a bottle over before any handoff. (Middle) After a handoff from a tilted pose, the policy only lifts the gripper above the alarm clock, and later presses tip the clock over. (Right) After self-improvement, the policy drops the fan during a whole-task call in the hard setting. The left and middle cases are failed C2R episodes of \mref{tab:main}.}
\label{fig:supp_failures}
\end{figure}

\paragraph{Failure cases.}
Our system fails in 8 of the 500 C2R episodes in \mref{tab:main}: four episodes reach the limit on language-model calls or time, the planner motions of the agent knock an object over in two, and a contact segment fails in the other two. In the left case of \cref{fig:supp_failures}, the arm knocks a bottle over with a planner motion before any handoff. In the middle case, the alarm clock lies beyond the reach of a top-down press, so the agent places the gripper in a tilted pose that differs from the start poses in $\mathcal{D}_0$. From this pose, the policy only lifts the gripper above the clock, and later presses tip the clock over. The right case comes from the evaluation after self-improvement and is not among these eight failures: during a whole-task call, the policy drops the fan.

\subsection{Deployment on a Real Robot}
\label{sec:supp_real}

\begin{figure}[!htb]
\centering
\includegraphics[width=\linewidth]{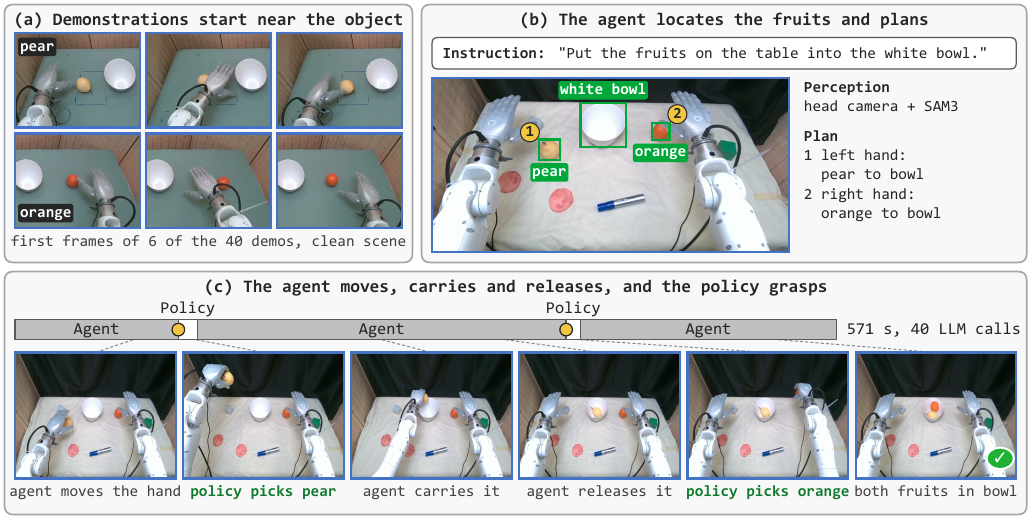}
\caption{\textbf{Deployment on a real dual-arm robot with dexterous hands.} (a) The demonstrations are collected in a clean scene and start from varied hand poses near each fruit. Each image is the first head-camera frame of one demonstration. (b) Given the instruction, the agent locates the fruits in the head-camera image and plans one hand for each fruit. (c) The agent hands over to the policy only to grasp each fruit and performs all other steps by itself.}
\label{fig:supp_real}
\end{figure}

\paragraph{Setup.}
We deploy our system on a real dual-arm robot with two dexterous hands, a head camera and a wrist camera on each arm. The task is to put a pear and an orange into a white bowl, and the instruction is ``Put the fruits on the table into the white bowl.'' For boundary-explicit training, we collect 20 teleoperated demonstrations of picking up each fruit in a clean scene, which contains only the two fruits and the bowl. As in simulation, each demonstration records only the interaction with the object, from a hand pose near the fruit until the fruit is lifted. These start poses vary around the fruit (\cref{fig:supp_real}a), so that the handoff boundary of the policy is wide. We fine-tune $\pi_{0.5}$ on the 40 demonstrations. The test scene differs from this clean scene, with several distractor objects and a different tablecloth (\cref{fig:supp_real}b). The agent is the same as in simulation, and its tools observe the scene, segment objects with SAM3, move a hand to a target pose, open a hand and call the policy.

\paragraph{Results.}
\Cref{fig:supp_real} shows one episode. The agent first reads the instruction, segments the head-camera image, and locates the pear on the left and the orange on the right (\cref{fig:supp_real}b). It then plans to pick up each fruit with the hand on the same side. For the pear, the agent moves the left hand to a start pose near the fruit and hands over to our policy, which grasps and lifts the pear (\cref{fig:supp_real}c). The agent then carries the pear above the bowl and opens the hand, and it repeats these steps for the orange with the right hand. Both fruits end in the bowl, and the task is completed. This first attempt covers a single task, and we leave a quantitative evaluation on real robots to future work.

\paragraph{Outlook.}
In early attempts, we let the agent grasp the fruits by itself with commands that close the hand, and none of these attempts succeeded. This suggests that the agent alone still has a long way to go on tasks with dexterous hands, and we attribute this to two reasons. First, a dexterous hand has many degrees of freedom, which are hard to coordinate through discrete commands. Second, the interaction between the fingers and the object is a problem of contact dynamics. The fingers must respond to the contact continuously, while the agent senses contact only indirectly and acts at a far lower rate (\cref{sec:supp_cl_weaknesses}). Handing the grasp to a learned policy compensates for part of this weakness. However, a policy that only grasps is not enough, as the value of a dexterous hand lies in adjusting the object within the hand during the task. What capabilities a learned policy should have in dexterous manipulation, and how the agent should steer it, are therefore directions worth exploring.

%% file: sections/supp/tab_robodojo_scores.tex
\begin{table}[!htb]
\centering
\caption{\textbf{Success rates and scores on RoboDojo.} The success rates repeat \mref{tab:main}, and Avg.\ is weighted by the number of tasks in each subset. $^\dagger$Reported by the official release, which gives no score for this subset (n/a). Best values are bold.}
\label{tab:supp_robodojo_scores}
\vspace{2pt}
\small
\setlength{\tabcolsep}{9pt}
\renewcommand{\arraystretch}{1.08}
\begin{tabular}{@{}lcccc@{}}
\toprule
\textbf{Method} & \textbf{Gen.} & \textbf{Mem.} & \textbf{Open} & \textbf{Avg.} \\
\midrule
\multicolumn{5}{@{}l}{\textit{Success rate (\%)}} \\
$\pi_{0.5}$~\cite{pi05_2025} & 16.7 & 3.3 & 0.0 & 8.5 \\
AGP~\cite{jia2026agp} & 10.0 & 46.7 & 50.0 & 30.8 \\
Harness VLA~\cite{harnessvla2026} & 41.7$^\dagger$ & 36.7 & 42.5 & 40.8 \\
Ours & \textbf{62.5} & \textbf{80.0} & \textbf{71.3} & \textbf{69.2} \\
\midrule
\multicolumn{5}{@{}l}{\textit{Score}} \\
$\pi_{0.5}$~\cite{pi05_2025} & 24.8 & 4.0 & 0.4 & 12.5 \\
AGP~\cite{jia2026agp} & 13.7 & 51.3 & 51.9 & 34.1 \\
Harness VLA~\cite{harnessvla2026} & n/a & 37.7 & 43.8 & n/a \\
Ours & \textbf{66.8} & \textbf{80.5} & \textbf{73.0} & \textbf{71.9} \\
\bottomrule
\end{tabular}
\end{table}

%% file: sections/supp/sec_F_prompt.tex
\section{Prompt and Memory}
\label{sec:supp_prompt}

This section shows the evaluation prompt of the agent after boundary-explicit training, which \mref{sec:setup} refers to, and two entries of its memory.
The prompt builds on that of Harness VLA~\cite{harnessvla2026}. \cref{tab:supp_prompt_protocol} shows the sections that we add, which contain the handoff protocol described in \cref{sec:supp_stage1}, and \cref{tab:supp_prompt_base} shows the sections that are kept from Harness VLA. The read order in \cref{tab:supp_prompt_base} points the agent to the task-specific recipe at the top of \cref{tab:supp_memory}.

\paragraph{Prompt structure and size.}
The handoff rules described in \cref{sec:supp_stage1} are embedded in two longer sections, on the local-skill handoff and on pose suggestions, which also explain the geometry needed to apply them, such as the frame conventions and the conversion between end-effector and fingertip poses. Two further sections describe how to call the policy and how to word its commands, and point to the list of the 278 distinct commands in $\mathcal{D}_0$. Together, these four sections take about 1{,}700 of the 2{,}600 words of the prompt (\cref{tab:supp_prompt_protocol}). The other sections are mostly kept from Harness VLA, including its loop of acting and then inspecting fresh observations. The agent also reads a pose guide of about 1{,}000 words that restates the pose suggestions. These sections and the guide contain no example images or task-specific descriptions.

\input{sections/supp/box_prompt_eval}

\input{sections/supp/box_memory}

%% file: sections/supp/box_prompt_eval.tex
\providecommand{\pnote}[1]{\colorbox{black!7}{\rmfamily\color{black!60}#1}}
\providecommand{\phead}[1]{\medskip\noindent\pnote{\textbf{\small #1}}\par\nopagebreak\smallskip}
\providecommand{\pph}[1]{{\rmfamily\itshape\color{black!55}$\langle$#1$\rangle$}}
\providecommand{\pus}{\textunderscore\allowbreak}
\providecommand{\pnotepar}[1]{\par\noindent\colorbox{black!7}{\parbox{\dimexpr\linewidth-2\fboxsep\relax}{\rmfamily\footnotesize\color{black!60}#1}}\par}
\par\ifdim\dimexpr\pagegoal-\pagetotal\relax<18\baselineskip\clearpage\fi
\captionof{table}{\textbf{Sections of the evaluation prompt that we add to the prompt of Harness VLA.} Besides the four sections of the handoff protocol, three sections describe the segmentation tool, the accessible files and the start of the episode. In the full prompt, the first section follows ACCURACY-FIRST LOOP of \cref{tab:supp_prompt_base}, the next five follow MODE, and BEGIN comes last. The prompt is instantiated for the \emph{adjust bottle} task in a randomized scene. File paths are replaced by placeholders in angle brackets, internal names are removed, and a few outdated phrases about the policy, such as the type of its actions, are updated to match the policy that we evaluate. Shaded headings are the section titles of the prompt itself.}
\label{tab:supp_prompt_protocol}
\nopagebreak
\begin{tcolorbox}[enhanced, breakable, colback=white, colframe=black!55, boxrule=0.5pt, arc=1pt,
 left=6pt, right=6pt, top=4pt, bottom=4pt, before skip=8pt, after skip=8pt,
 title after break={Tab.~\ref{tab:supp_prompt_protocol} (continued)}, colbacktitle=black!7, coltitle=black!70,
 fonttitle=\footnotesize\rmfamily\bfseries, toptitle=1pt, bottomtitle=1pt]
\footnotesize\ttfamily\raggedright\frenchspacing\setlength{\parskip}{3pt}\widowpenalty=10000\clubpenalty=10000
\phead{VLA AND PRIMITIVE CONTROL}

The only available VLA is the contact-segment $\pi$0.5 policy, trained only on self-collected data with three wide cameras, invoked with pi05\pus{}act(prompt=..., chunks=1 to 4). Choose one visually grounded subtask instruction resembling the training references below. The prompt is actually passed to $\pi$0.5. Describe the intended arm, target, and operation clearly; do not invent hidden scene facts. One chunk is 50 absolute world TCP pose and gripper-command actions at 40ms each: 2 seconds of simulation. Multiple chunks re-observe and replan with $\pi$0.5 between chunks; you receive the latest view when the tool call returns. Choose the call length using the optional guidance below, then inspect actual progress. Use the other primitives for visible geometric staging when appropriate. There is no second VLA or alternate VLA checkpoint in this session.

\phead{SAM3 VISUAL LOCALIZATION}

Use segment(text\pus{}prompt="bottle", view="head", step=...) to locate a visible object with official SAM3. Available views are head, left\pus{}wrist, right\pus{}wrist, all wide cameras. Query a concise visible object description, including distinguishing color or part when useful. The tool segments the saved RGB observation at the specified step (latest by default); it does not move the robot. It returns found, score, box\pus{}px {[}x\pus{}min,y\pus{}min,x\pus{}max,y\pus{}max{]}, mask\pus{}center\pus{}rc {[}row,col{]}, and mask\pus{}area\pus{}px. A mask center is an image location, not a grasp pose, and may lie on background for hollow objects. Inspect the image and choose a visible surface/\allowbreak{}grasp region. Query sample\pus{}world\pus{}xyz with pixels in {[}row,col{]} order at the SAME view and step to obtain depth coordinates; never treat pixel coordinates as world coordinates. Re-observe and re-query after motion. If found=false or a mask is ambiguous, use another view/\allowbreak{}description or visually localize; do not invent a detection. A detection does not certify a grasp or task success.\\
Geometric move\pus{}to uses native speed-limited TOPP execution. Check achieved EEF pose and final\pus{}dist\pus{}m; execution success does not guarantee arrival through an obstruction. pi05\pus{}act alone executes each policy command for 40ms.

\phead{DEFAULT LOCAL-SKILL HANDOFF}

\textbf{Hybrid planner /\allowbreak{} primitive /\allowbreak{} VLA framework}\\*
You operate a hybrid framework: Claude plans and observes, free-space primitives support approach and staging, and pi05\pus{}act executes learned local manipulation skills. Prefer the available contact-segment $\pi$0.5 VLA for suitable skill segments such as grasping and local object manipulation, using a concise grounded subtask instruction. Do not default to manually scripting a suitable learned skill through repeated move\pus{}to/\allowbreak{}gripper calls. Use primitives where they help with approach, clearance, transport or recovery, then hand the suitable skill segment to the VLA and inspect the result. Select tools from the current scene and progress; VLA preference does not require an immediate opening call from a distant home pose or repeated calls without checking progress.

\textbf{Suggested position-and-orientation handoff to the local-skill VLA}\\*
Inspect the CURRENT head wide and both wrist wide images and measured EEF/\allowbreak{}TCP poses. Identify the target, active arm, skill phase, approach direction and jaw-closing direction. From a distant empty-hand home pose, a useful default is to choose a grasp-compatible orientation, stage with clearance, inspect the achieved orientation AND position, then use pi05\pus{}act for the local pick. An EEF or TCP located above the object can still point horizontally; position alignment alone does not establish a suitable starting pose.

The distances, angular ranges and example quaternions below are suggestions, not admission gates, exact handoff start pose requirements, reasons to reset, or limits on where the policy can work. Use current geometry to adapt them. If already suitably positioned and oriented, a direct VLA call can be appropriate. If holding an object, preserve the verified grasp and stage that continuation rather than returning to an empty-hand pickup pose. A concise action note describing the intended approach, jaw alignment and TCP offset can help check your choice before moving.

\phead{PI05 TRAINING INSTRUCTION REFERENCE}

\textbf{Contact-segment $\pi$0.5 trained-style instruction vocabulary}\\*
Use one concise English subtask instruction: operation + visible target + active arm. For coordinated manipulation, describe each arm's role. Ground target identity, relative relations and arm choice in current observations. Avoid long chains of unrelated goals and numerical joint commands as language. Exact training-label wording is available in \pph{guide directory}/\allowbreak{}PI05\pus{}TRAINING\pus{}INSTRUCTIONS.md; read the relevant entries as vocabulary references, not task plans, object-presence claims or guarantees of capability. New object/\allowbreak{}operation/\allowbreak{}arm combinations are generalization. The guide contains 278 instruction phrases from the self-collected training subset (2699 episodes, excluding its validation split). It is a wording reference, not an exhaustive inventory.

\phead{PI05 TEXT POSE SUGGESTIONS}

\textbf{Text pose suggestions: grasp direction, jaw direction and distance together}\\*
Use live RGB/\allowbreak{}depth and measured robot poses to estimate geometry; no prior training-example pictures or live target coordinates are supplied. These are operational suggestions, not compulsory call preconditions. Favor a compatible local start while allowing a justified alternative when the scene or held-object state calls for one.

\textbf{Robot frame and quaternion meaning}\\*
move\pus{}to targets the EEF in world metres, with quaternion {[}qw,qx,qy,qz{]}, normalized to unit length. The current ALOHA-AgileX interface's local +X points from EEF toward TCP/\allowbreak{}fingertips; local +/\allowbreak{}-Y is the jaw opening/\allowbreak{}closing axis. World +Z is upward. For a downward pickup, a useful suggestion is local +X approximately toward world -Z, often within about 15 degrees when unobstructed. Keep the jaw-closing axis approximately horizontal and across the object's narrow grasp dimension. For an elongated grasp region, align the jaw axis approximately perpendicular to its visible long axis. For a box-shaped grasp region, close across a suitable pair of opposing faces. Choose yaw from visible geometry; about +/\allowbreak{}-15 degrees around a suitable grasp yaw is a useful local variation, not an absolute yaw range.

\textbf{Concrete orientation examples (wxyz; suggestions, not universal poses)}\\*
{[}0.70710678,0,0.70710678,0{]} sends local +X downward and jaw axis +Y along world +Y. This can suit a horizontal handle whose long axis is approximately world X. {[}0.5,-0.5,0.5,0.5{]} also sends local +X downward, but its jaw axis is along world -X; it can suit a handle approximately along world Y. Both approach downward, yet their jaw alignment differs by 90 degrees. For a handle at table-plane angle alpha from world X, an illustrative downward orientation is q\pus{}worldZ(alpha) multiplied on the LEFT of {[}0.70710678,0,0.70710678,0{]}; select the feasible yaw equivalent matching the visible jaw/\allowbreak{}object relation. q and -q represent the same rotation.

Do not assume a numerical quaternion points downward just because it is nonzero or was explicitly passed. In this interface {[}0.7,0,0,0.714{]} is close to the initial home orientation: its local +X is approximately horizontal toward world +Y. Keeping that orientation while raising z moves a horizontal gripper above the object; it does not rotate the fingers downward. Omitting quat in move\pus{}to preserves the CURRENT EEF orientation. That is useful for intentional orientation-preserving transport, but it does not perform an implicit grasp-orientation alignment. When choosing a different approach, consider passing an explicit compatible quat and checking the achieved pose.

Useful checks from q=(w,x,y,z): local +X in world is {[}1-2(y*y+z*z), 2(x*y+w*z), 2(x*z-w*y){]}. The jaw axis local +Y in world is {[}2(x*y-w*z), 1-2(x*x+z*z), 2(y*z+w*x){]}. Compare these directions with world down and the visible object axis, not just the xyz error. These checks are planning aids, not runtime gates. Re-observe the real achieved pose because a reachable plan or tool success does not certify visual alignment or a grasp.

\textbf{Free-hand pickup approach and posture suggestions}\\*
Before handing an empty-hand pickup to $\pi$0.5, use the head view to identify the target, a suitable grasp region and nearby obstructions. Combine both wrist views to judge the distance, approach direction and the placement of both fingers. Prefer a region the fingers can enclose from opposite sides with stable contact; aiming at the visual center alone may not give a secure grasp.

For an object suited to an overhead pickup, a useful starting suggestion is to point the fingertips toward the grasp region and keep the TCP/\allowbreak{}fingertip grasp center roughly 8-12 cm above it. Center the open jaws over the part you intend to grasp. Judge this distance from the fingertip grasp center, not the wrist. Being above the object is insufficient: check that the fingertips face the object and that closing the jaws would enclose the intended region from its two sides.

Adapt the jaw direction to the visible shape. For a handle, rod or other elongated region, let the fingers close across its narrower sides rather than along its length where they may miss. For a block-like object, choose opposing surfaces with stable, sufficient contact area. There is no universal grasp orientation: both the approach direction and closing direction should suit the current grasp region.

When approaching empty-handed, opening the gripper enough to accommodate that region with some clearance can help. In the wrist views, check whether the fingers lie on opposite sides of the target, whether either finger is obstructed, and whether further approach would first hit the table or a neighboring object. If needed, make a small adjustment and inspect the achieved pose rather than chaining blind movements.

For occlusion, restricted clearance or shapes unsuitable for overhead pickup, adapt the approach height, direction and jaw orientation instead of forcing an overhead entry. If already in a suitable local grasp pose, directly invoking $\pi$0.5 can be appropriate without an extra staging move.

These are suggestions for choosing a useful start, not compulsory call conditions. If an object is already stably held, preserve that grasp and stage the next phase; do not reopen the gripper or return to an empty-hand pickup pose.

\textbf{EEF/\allowbreak{}TCP conversion during staging}\\*
For this robot TCP = EEF + R(quat)*{[}0.12,0,0{]} metres; confirm against the current observed EEF/\allowbreak{}TCP poses. At a DOWNWARD orientation, a chosen TCP goal normally requires an EEF about 12 cm ABOVE that TCP goal. For a changed orientation use EEF\pus{}goal = desired\pus{}TCP - R(desired\pus{}quat)*{[}0.12,0,0{]}. The simpler translation EEF\pus{}goal=current\pus{}EEF+(desired\pus{}TCP-current\pus{}TCP) is appropriate only when keeping the current orientation. When changing orientation, a clear high staging pose followed by re-observation can be convenient; a wrist rotation changes the TCP position even if EEF xyz is held constant. Do not carry a horizontal-pose TCP offset into a downward-pose move.

\textbf{Held-object continuation suggestions}\\*
Verify that an object moves with the gripper before treating it as held. Preserve a stable grasp and account for the held object's geometry during staging and transport. Judge the relevant object surface or contact point against its destination, rather than using wrist proximity alone. Consider approach direction, relative surface orientation, clearance and lateral alignment together. Adapt clearance to current geometry and uncertainty; inspect contact, support and release where relevant. Preserve already achieved relations and any occupied arm's useful role. Do not force a held-object continuation back into an empty-hand pickup pose.

\textbf{Call length and completion}\\*
One chunk is 50 absolute world TCP pose and gripper-command actions at 40ms each = 2 simulated seconds. Clear pickups may use 3-4 chunks; precision contact or a held-object continuation may start with 1-2 and inspection. Choose duration from visible progress. Tool success means execution; TASK\pus{}ENV.eval\pus{}success is the benchmark signal. Describe actual visual grasp, contact, orientation and support separately rather than inferring them from a closed gripper or the benchmark signal alone.

The free-pick local skill offsets commonly around 8-12 cm and yaw variation +/\allowbreak{}-15 degrees are prior training-derived suggestions. These are skill-reference offsets, not live object coordinates. Frame-axis and 12 cm EEF/\allowbreak{}TCP facts come from the ALOHA-AgileX robot model/\allowbreak{}interface. The tilt and general continuation advice here are practical suggestions, not additional measured training statistics.

\phead{DEDICATED PLANNER WORKSPACE}

Only the curated operating guide and instruction-vocabulary guide, the selected memory corpus, and this instance's own audit/\allowbreak{}attempt notes are accessible. Do not search source code, training datasets, runtime logs, other runs, personal workspaces, shell or external services. File tools enforce this allowlist; a denied path is not a reason to try alternate paths. Write only the current audit, failed-attempt records and your current exploration inbox.

\phead{BEGIN}

Read the guide and this instance's memory, then inspect current head/\allowbreak{}left-wrist/\allowbreak{}right-wrist wide images and measured EEF/\allowbreak{}TCP poses. Use the text pose suggestions to consider position, approach angle, jaw-closing direction and distance together. From an empty distant home pose, default to a deliberate compatible approach; being above or close to the target alone does not establish a matching grasp orientation. If already suitable or holding the object for a continuation, direct pi05\pus{}act use can be appropriate. All pose examples and ranges are suggestions, not exact handoff start pose requirements or mandatory admission checks. Inspect achieved live pose and progress. Only the contact-segment $\pi$0.5 policy is available as a VLA.
\end{tcolorbox}

\par\ifdim\dimexpr\pagegoal-\pagetotal\relax<14\baselineskip\clearpage\fi
\captionof{table}{\textbf{Sections of the evaluation prompt that are kept from Harness VLA.} Their text is that of the official RoboTwin~2.0 prompt of Harness VLA, except for the name of the policy and for steps 1 and 2 of READ ORDER, which point to our pose guide and pose suggestions. From the official prompt, we remove the section of task-family playbooks and replace the section on policy control by the one in \cref{tab:supp_prompt_protocol}. Placeholders and headings are as in \cref{tab:supp_prompt_protocol}.}
\label{tab:supp_prompt_base}
\nopagebreak
\begin{tcolorbox}[enhanced, breakable, colback=white, colframe=black!55, boxrule=0.5pt, arc=1pt,
 left=6pt, right=6pt, top=4pt, bottom=4pt, before skip=8pt, after skip=8pt,
 title after break={Tab.~\ref{tab:supp_prompt_base} (continued)}, colbacktitle=black!7, coltitle=black!70,
 fonttitle=\footnotesize\rmfamily\bfseries, toptitle=1pt, bottomtitle=1pt]
\footnotesize\ttfamily\raggedright\frenchspacing\setlength{\parskip}{3pt}\widowpenalty=10000\clubpenalty=10000
\phead{ROLE}

You control one dual-arm RoboTwin demo\pus{}randomized episode through the registered tools. Satisfy the complete current task\pus{}language in one no-restart episode. Prefer one accurate, recipe-supported sequence over broad exploration, and protect every achieved subgoal.

\phead{READ ORDER}

Before the first robot mutation:\\
1. Read \pph{guide directory}/\allowbreak{}GUIDE\pus{}PI05\pus{}POSE.md completely.\\
2. Inspect view\pus{}env\pus{}state(step=0), its CURRENT head wide and both wrist wide images, and measured EEF/\allowbreak{}TCP position AND orientation. Read the text pose suggestions.\\
3. Read \pph{memory directory}/\allowbreak{}task-specific/\allowbreak{}robotwin\pus{}adjust\pus{}bottle\pus{}demo\pus{}clean\pus{}s0.json and \pph{memory directory}/\allowbreak{}task-specific/\allowbreak{}robotwin\pus{}adjust\pus{}bottle\pus{}demo\pus{}clean\pus{}s0\pus{}recipe.jsonl when present.\\
4. Read \pph{memory directory}/\allowbreak{}MEMORY.md and at most one to three relevant leaves.

The current task\pus{}language and fresh observation override historical memory. Use the semantic JSON as the phase plan and the JSONL as evidence for action type and VLA cadence, never as a coordinate replay.

\phead{CLEAN-TO-RANDOMIZED TRANSFER}

Transfer roles, phase order, required arms, observable gates, VLA/\allowbreak{}analytic division, chunk pattern, terminal action, and known failures. Rebind every object, destination, relation, arm choice, pixel, pose, table height, clearance, grasp point, and release/\allowbreak{}contact point from this episode. Use a supported recipe as the default skeleton when current evidence agrees; treat an experimental recipe as a weak prior.

\phead{ACCURACY-FIRST LOOP}

Issue one registered action, inspect fresh before/\allowbreak{}after evidence, then decide again. Maintain a compact internal ledger: current phase, achieved/\allowbreak{}protected relations, held object and arm, first unmet postcondition, blocker, and next observable gate. Advance only when the current gate is visibly satisfied. Primitive success is not task success.

If an action makes useful progress but stops mid-phase, continue the same phase with the shortest suitable action. The contact-segment $\pi$0.5 policy may be called repeatedly as the recipe and physical state require; lack of an immediate completed gate does not by itself forbid another VLA chunk. The two-no-progress rule applies to an unchanged analytic primitive target or identical hand-written recovery: after two ineffective repetitions, re-observe and change one meaningful variable. Near success, repair only the remaining blocker; do not restart the full task or disturb correct objects.

\phead{PERCEPTION}

Use the head view as semantic authority for identity, distractors, destinations, language relations, and global progress. Use the matching current wrist view to refine geometry for that same chosen candidate; do not let it silently switch to a look-alike. Pair RGB and world maps from the same step, view, and resolution. World maps are {[}row,col{]} -\textgreater{} {[}x,y,z{]} metres and may contain NaN; visible surface points are not automatically object centers. Relocalize after occlusion, contact, or substantial arm/\allowbreak{}object motion.

\phead{RUNTIME}

The registered RoboTwin Toolkit is the only control surface. Do not use shell, Python, network clients, legacy command files, plan mode, user questions, or unrelated built-in tools. Never inspect task source, evaluator implementation, hidden rewards, object poses, raw expert trajectories, another attempt, or unapproved historical geometry. The curated files under \pph{memory directory} are approved planning references and are not subject to this restriction. Call the selected registered tool in the same response instead of announcing a future action. The episode is non-interactive and must not be restarted.

\phead{BUDGET AND SUCCESS}

Track remaining\pus{}steps = step\pus{}lim - take\pus{}action\pus{}cnt. The 10000 native-step limit is a safety ceiling, not a target. The same-task recipe and its phase count are the soft complexity prior: short tasks should usually stay concise; long ranking, stacking, container, or articulated tasks may need more phases. Extra budget never justifies repeating an ineffective strategy. Also preserve enough Planner turns and wall time to verify and finish.

Only fresh TASK\pus{}ENV.eval\pus{}success=true confirms success. Stop robot actions immediately after native success or budget exhaustion. Every exit must call finish exactly once after a fresh status check, reporting failure honestly when native success remains false.

\phead{MODE}

Solve the current episode now using registered tools and current evidence. Do not ask for clarification or defer the next determined action.

\phead{CELL}

- task: adjust\pus{}bottle\\
- seed: 100000\\
- task\pus{}config: demo\pus{}randomized\\
- checkpoint: the contact-segment $\pi$0.5 policy
\end{tcolorbox}

%% file: sections/supp/box_memory.tex
\providecommand{\pnote}[1]{\colorbox{black!7}{\rmfamily\color{black!60}#1}}
\providecommand{\phead}[1]{\medskip\noindent\pnote{\textbf{\small #1}}\par\nopagebreak\smallskip}
\providecommand{\pph}[1]{{\rmfamily\itshape\color{black!55}$\langle$#1$\rangle$}}
\providecommand{\pus}{\textunderscore\allowbreak}
\providecommand{\pnotepar}[1]{\par\noindent\colorbox{black!7}{\parbox{\dimexpr\linewidth-2\fboxsep\relax}{\rmfamily\footnotesize\color{black!60}#1}}\par}
\par\vspace{-8pt}\noindent%
\begin{minipage}{\linewidth}
\captionof{table}{\textbf{Two memory entries read by the agent.} Top: the task-specific recipe of \emph{adjust bottle}, which is recorded during exploration in the clean scene and read at evaluation in randomized scenes. Bottom: a global failure note written during exploration on two other tasks, shown without its header. Shaded text is our annotation, and one phrase is updated to match the action space of the policy.}
\label{tab:supp_memory}
\vspace{-5pt}%
\begin{tcolorbox}[enhanced, colback=white, colframe=black!55, boxrule=0.5pt, arc=1pt,
 left=6pt, right=6pt, top=4pt, bottom=4pt, before skip=2pt, after skip=0pt]%
\footnotesize\ttfamily\raggedright\frenchspacing\setlength{\parskip}{3pt}\widowpenalty=10000\clubpenalty=10000
\phead{Task-specific recipe: \pph{memory directory}/task-specific/robotwin\_adjust\_bottle\_demo\_clean\_s0\_recipe.jsonl}

\pnotepar{Two JSONL lines, one per state-changing command of the successful exploration in the clean scene. Line breaks and indentation are ours.}

\{"action": "move\pus{}to", "arm": "left",\\
\hspace*{1em}"xyz": [-0.04, -0.165, 1.02],\\
\hspace*{1em}"quat": [0.70105, 0.0923, 0.70105, -0.0923],\\
\hspace*{1em}"gripper": 1\}\\
\{"action": "pi05\pus{}act",\\
\hspace*{1em}"prompt": "Turn the bottle upright with the left arm.",\\
\hspace*{1em}"chunks": 3\}

\phead{Global failure note: \pph{memory directory}/global/global\_failure\_pi05\_moves\_idle\_arm.md}

\pnotepar{Written by an exploration agent and merged into the frozen memory. The header of the note is omitted.}

$\pi$0.5 outputs absolute end-effector pose actions for both arms. During a right-arm pick, the parked left arm drifted several cm and its gripper closed from 1.0 to about 0.03.

What to do:\\
- After any pi05\pus{}act call, check the idle arm's pose and gripper.\\
- Before expecting task success, release (open) any idle gripper and keep the idle arm clear of placed objects.\\
- In place\pus{}bread\pus{}basket, native success fired only once the idle left gripper was reopened.
\end{tcolorbox}
\end{minipage}
\par

%% file: sections/supp/sec_closer_look.tex
\providecommand{\clfig}{fig/supp}%
\providecommand{\cltab}{sections/supp}%
\providecommand{\cltex}{sections/supp}%
\input{\cltab/tab_cl_combined}
\clearpage
\section{A Closer Look at Agent-Direct Control}
\label{sec:supp_closer_look}

This section reports the study behind a statement in \mref{sec:intro}: an agent built on a general-purpose multimodal model is strong at task-level decisions and generalizes across layouts and instructions, but it executes contact poorly. Before building our system, we let such an agent control a dual-arm robot directly on RoboTwin~2.0 and RoboDojo without any robot training, which we call agent-direct control, and we re-examine every failed trial. \cref{sec:supp_cl_setup} describes the setup, \cref{sec:supp_cl_strengths} and \cref{sec:supp_cl_weaknesses} report what the agent does well and where it falls short, and \cref{sec:supp_cl_implications} discusses the research directions that follow.

\subsection{Setup}
\label{sec:supp_cl_setup}

\paragraph{Agent and interface.}
The agent is Claude Opus 5.5 with xhigh reasoning effort, the level between the high and max efforts in \mref{tab:ablation_llm}, and we give it no robot training data, no demonstrations and no memory between trials. It observes RGB images from the wrist cameras and from fixed cameras with known poses, together with the proprioceptive state, but no depth, table height or object poses. It acts through 11 low-level commands, such as capturing images, reading the robot state, moving an end effector or the joints and opening or closing a gripper, and the command counts below include only those that capture images or move the robot. No skill is provided, so the agent writes its own scripts for perception and geometry at run time. The prompt contains only general rules, such as checking images before and after each action, logging the expected and the actual outcome of each motion and continuing from the current state after a failure, and no task-specific procedure.

\begin{figure}[!htb]%
\centering
\includegraphics[width=\linewidth]{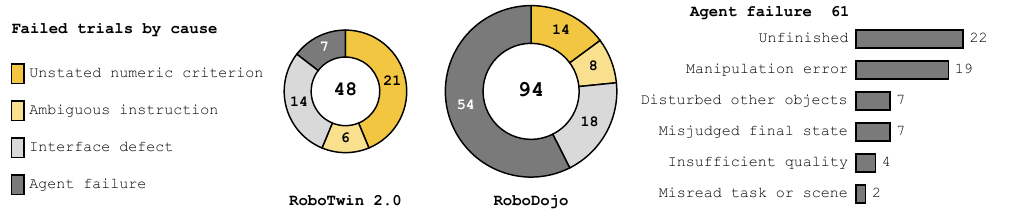}
\caption{\textbf{Fewer than half of the failed trials are attributed to the agent itself.} Causes of the 142 failed trials from a post-hoc review. Ring areas are proportional to the numbers of failed trials, and the bars divide the 61 agent failures into six types.}
\label{fig:supp_cl_causes}
\end{figure}

\begin{figure}[t]
\centering
\includegraphics[width=\linewidth]{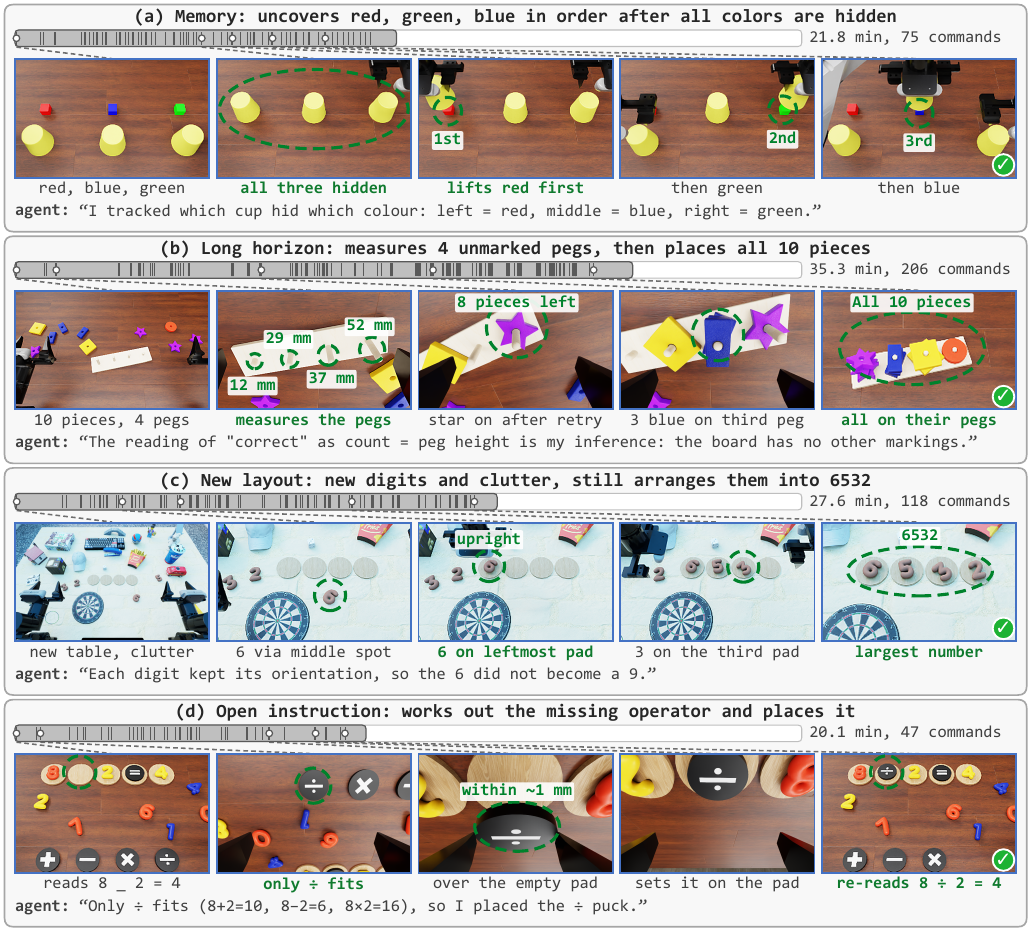}
\caption{\textbf{Without robot training, the agent handles memory, long horizons, new layouts and open instructions.} Each row shows one trial. (a) Memory: asked to cover three blocks and then uncover them in a given order, the agent tracks which cup hides which block. (b) Long horizon: asked to place 10 pieces on the correct pegs, the agent measures the four unmarked pegs and matches each peg height to a number of pieces. (c) New layout: asked to form the largest number, the agent arranges the digits into 6532 on a new table with clutter. (d) Open instruction: asked to complete an equation, the agent checks each operator and places $\div$. The bar above each row shows the trial duration within the 45-minute limit, thin lines mark the commands, and circles mark the times of the frames. Labels on the frames come from the agent's own records, the last line quotes the agent verbatim, and the badge is the result of the benchmark's success check.}
\label{fig:supp_cl_abilities}
\end{figure}

\paragraph{Benchmarks and protocol.}
The study covers all 50 tasks of RoboTwin~2.0~\cite{robotwin2_2025} in clean scenes and 54 configurations of RoboDojo~\cite{chen2026robodojo}, namely its 42 tasks with both a standard and a randomized layout for each of the 12 generalization tasks. Each configuration gets one attempt of at most 45 minutes, and a failed configuration gets up to three attempts in total, which gives 204 trials (\cref{tab:supp_cl_combined}), 142 of which fail (\cref{fig:supp_cl_causes}). The benchmark's own success check judges the final state of each trial without returning its result to the agent. The study is exploratory, with one model and one scene per configuration, so it shows what the agent can do and how it fails, but not how reliably, and the trials in \cref{fig:supp_cl_abilities,fig:supp_cl_strategies,fig:supp_cl_failures} are selected examples. It is also earlier and simpler than our evaluation of AGP~\cite{jia2026agp} in \mref{tab:main}: it has the same kind of interface, but it repeats failed configurations and uses a higher reasoning effort, so the two sets of success rates are not directly comparable.

\def\cltabpos{!htb}%
\cltableresults

\newpage

\subsection{What the Agent Does Well}
\label{sec:supp_cl_strengths}

\paragraph{The agent is strong at task-level decisions without robot training.}
As shown in \cref{tab:supp_cl_combined}, the agent passes 72.0\% of the RoboTwin~2.0 tasks and 48.1\% of the RoboDojo configurations within three attempts. On RoboDojo, it passes 62.5\% of the open-vocabulary configurations on the first attempt but none of the precision configurations. In the four successful trials of \cref{fig:supp_cl_abilities}, the agent uncovers covered blocks in the instructed order from its own records, places 10 pieces on unmarked pegs, arranges digits into the largest number on a new table and works out the missing operator of an equation. These results suggest that the agent can generalize across layouts and instructions without robot training.

\begin{figure}[t]
\centering
\includegraphics[width=\linewidth]{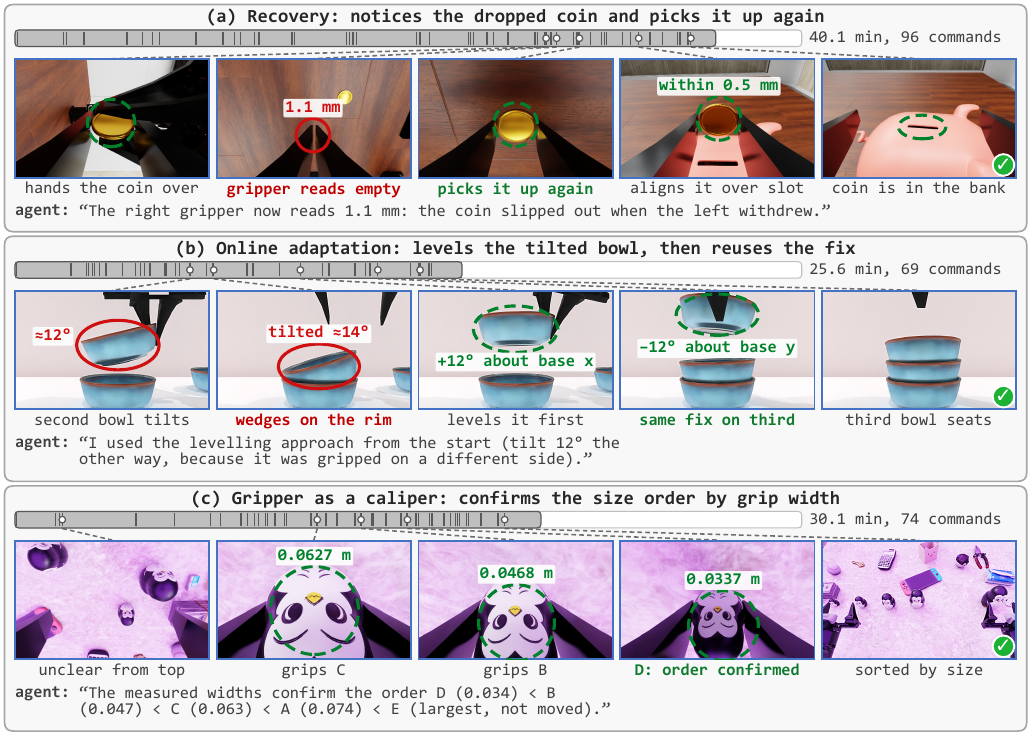}
\caption{\textbf{The agent devises the specific strategy for each task at run time.} Each row shows one trial. (a) Recovery: asked to insert a coin into a coin bank, the agent finds the coin after it slips out of the gripper, picks it up again and presses it into the slot. (b) Online adaptation: asked to stack three bowls, the agent levels a bowl that hangs tilted and wedges on the rim, and applies the same correction to the next bowl. (c) Gripper as a caliper: asked to sort five nesting dolls by size, the agent confirms their order from the gripper width after each grasp. Bars, labels, quotes and badges are drawn as in \cref{fig:supp_cl_abilities}.}
\label{fig:supp_cl_strategies}
\end{figure}

\paragraph{The agent devises the specific strategy for each task at run time.}
The prompt in \cref{fig:supp_cl_prompt} tells the agent to observe again, to change its approach when one keeps failing and to continue from the current state, but it gives no task-specific strategy. In \cref{fig:supp_cl_strategies}a, the gripper reads 1.1\,mm after a handover between the arms, so the agent infers that the coin has slipped out, finds it and presses it into the slot. In \cref{fig:supp_cl_strategies}b, after a bowl hangs tilted and wedges on the rim, the agent levels it by tilting the gripper the other way and applies the same correction to the next bowl. In \cref{fig:supp_cl_strategies}c, the agent confirms the size order of the dolls from the gripper width after each grasp. In another trial on RoboTwin~2.0, the agent fits its own kinematic model from joint samples and solves inverse kinematics with its own script. None of these strategies appears in the prompt, which suggests that the agent derives them from general rules at run time.

\subsection{Where the Agent Falls Short}
\label{sec:supp_cl_weaknesses}

\paragraph{Fewer than half of the failed trials are attributed to the agent itself.}
We re-examine every failed trial post hoc with a written rubric, against the final state in simulation and the source code of the benchmark's success check, neither of which the agent sees. As shown in \cref{fig:supp_cl_causes}, this review attributes 61 of the 142 failed trials to the agent itself and the rest to three other causes. \textbf{First}, with an unstated numeric criterion, the success check requires a threshold or a target that the instruction does not state. \textbf{Second}, with an ambiguous instruction, the success check reads the instruction differently from the agent. \textbf{Third}, interface defects come from the interface that we built around each benchmark, such as a reach limit that we set. The first two causes involve our judgment of what an instruction asks for, so we report all causes separately and do not correct the success rates in \cref{tab:supp_cl_combined}.

\begin{figure}[t]
\centering
\includegraphics[width=\linewidth]{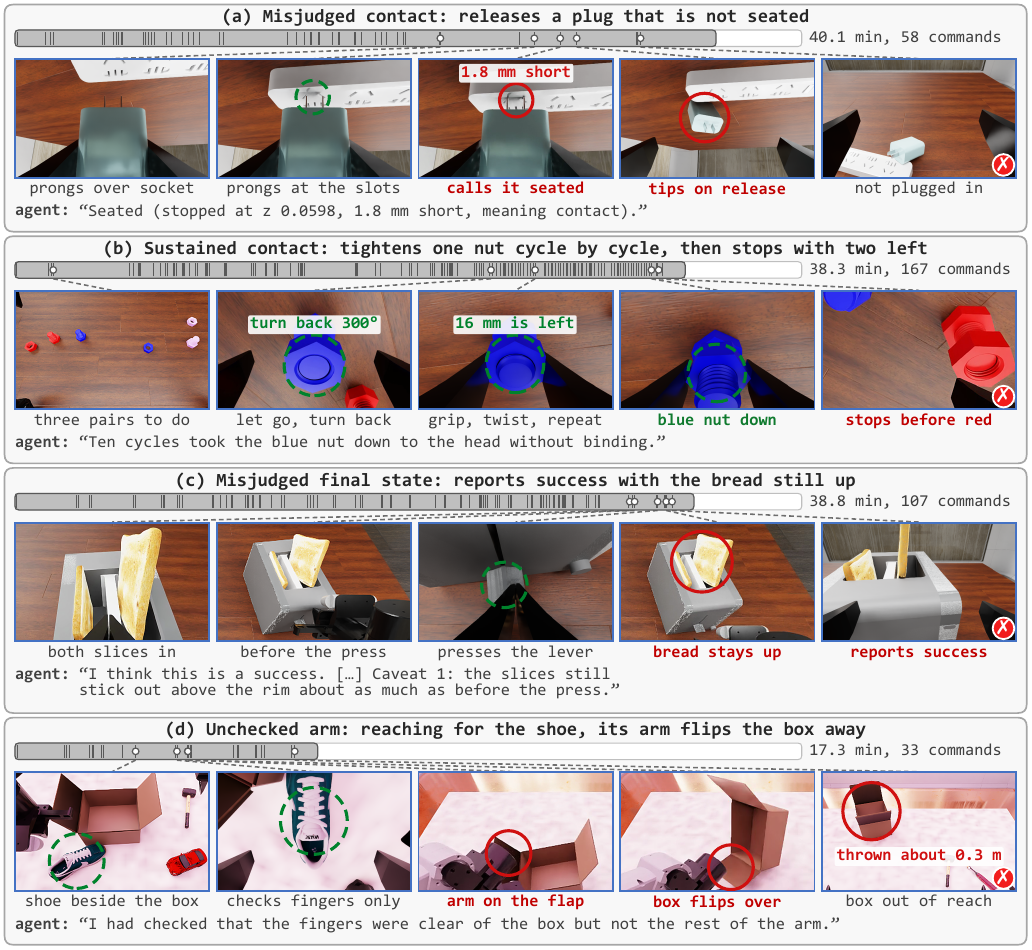}
\caption{\textbf{The agent fails mostly in physical interaction with objects.} Each row shows one failed trial. (a) Misjudged contact: asked to plug a charger into a power strip, the agent takes a stop 1.8\,mm short as contact and releases the plug before it is seated. (b) Sustained contact: asked to tighten three screws into their nuts, the agent finishes one and ends the trial before the other two are tightened. (c) Misjudged final state: asked to put two slices of bread into a toaster and press the lever down, the agent reports success although the bread has not gone down. (d) Unchecked arm: asked to place objects into a box, the agent checks its fingers but not the rest of its arm, which appears to press down a flap and throws the box about 0.3\,m away. Bars, labels, quotes and badges are drawn as in \cref{fig:supp_cl_abilities}.}
\label{fig:supp_cl_failures}
\end{figure}

\paragraph{The failures of the agent itself mostly lie in physical interaction with objects.}
As shown in \cref{fig:supp_cl_causes}, most failures attributed to the agent occur on RoboDojo, in tasks such as hanging mugs, inserting a plug and fastening screws, and most of them are unfinished tasks or manipulation errors. In the four failed trials of \cref{fig:supp_cl_failures}, the agent releases a plug that it wrongly judges seated, ends the trial with one of three nuts tightened, reports success although the bread has not gone down, and knocks a box away with an unchecked part of its arm. Two explanations remain to be tested: the agent spends about half a minute per command, and it senses contact only indirectly, through images and the gripper opening. Both would mostly affect physical interaction with objects, which is the part that the system in this paper hands over to a learned policy.

\paragraph{Most of the time goes to language-model inference.}
As shown in \cref{tab:supp_cl_time}, a trial takes a median of 18.2 minutes on RoboTwin~2.0 and 32.8 minutes on RoboDojo, whereas the median robot motion time is under one minute. Language-model inference takes 93.0\% of the total time on RoboTwin~2.0 and 86.5\% on RoboDojo, since the agent reasons over new observations between commands, and this time counts toward the time limit of a trial. Boundary-expanding self-improvement in \mref{sec:expandexp} reduces the language-model calls of our system, since the agent hands over to the retrained policy earlier.

\cltabletime%

\subsection{Discussion and Future Directions}
\label{sec:supp_cl_implications}

Agent-direct control replaces robot data with reasoning at run time. In our study, each strength of this choice comes with a cost, and each pair raises a research direction.

\paragraph{Agent-direct control is general but not dexterous.}
Reasoning lets the agent solve new tasks without training but does not make physical interaction with objects precise, which raises the question of how to combine reasoning with learned control. The missing ability can be trained into the language model, built into perception and motion tools, or handed over to a learned policy. The third route must say where the boundary between the agent and the policy lies and how the agent knows it.

\paragraph{Agent-direct control is deliberate but slow.}
Observing and reasoning between commands lets the agent recover and check its own work, but it also takes most of the time of a trial. In simulation, the scene waits while the agent reasons, whereas a real scene may keep changing. This raises the question of where reasoning should be spent, that is, which decisions need the language model and which do not.

\paragraph{Agent-direct control needs no training but does not learn from experience by itself.}
In our study, the agent keeps no memory between trials, so it solves each task as if for the first time. This raises the question of how experience becomes ability: whether it belongs in a memory, in reusable programs or in the weights of a policy, and whether the agent can become a source of robot data. Experience can also be trained into the agent itself: just as general agents have internalized computer use through large-scale post-training with reinforcement learning, we believe that they can internalize robot control from their experience on robots.

\paragraph{Agent-direct control follows language, but existing benchmarks do not evaluate it well.}
The two benchmarks in this study, like most existing ones, are designed for policies that learn from demonstrations, and our study meets four gaps when they evaluate an agent. \textbf{First}, their success checks contain conventions that demonstrations convey but the instruction does not state, as the unstated numeric criteria in \cref{fig:supp_cl_causes} show. \textbf{Second}, the simulation waits while the agent reasons, and the benchmarks count neither time nor cost. \textbf{Third}, a binary success signal cannot separate an instruction that the agent misunderstands from one that it fails to execute. \textbf{Fourth}, the benchmarks do not specify how many attempts the agent may make, what it may remember or whether it sees the success signal. Benchmarks designed for agentic robot manipulation may close these gaps by stating the required final state in the instruction and by fixing the time limit and the memory rules.

\medskip\noindent
Among the four research directions above, this paper mainly addresses the first, how to combine reasoning with learned control, and hands physical interaction with objects over to a learned policy. Its key idea is to define and expand the handoff boundary between the agent and the policy through training (\mref{sec:aligned_learning} and \mref{sec:expand}). In summary, a general agent already provides the task-level decisions of manipulation without robot training, and the open problems are how to combine it with learned control, where to spend its reasoning, how to turn its experience into ability and how to evaluate it. We believe that agents for robotics are a promising path toward general physical intelligence.

\input{\cltex/box_cl_prompt}

%% file: sections/supp/tab_cl_combined.tex
\ifdefined\clboxa\else\newsavebox{\clboxa}\newsavebox{\clboxb}\newsavebox{\clboxc}\fi
\providecommand{\cltabpos}{t}%
\def\cltableresults{%
\edef\cltabbegin{\noexpand\begin{table}[\cltabpos]}\cltabbegin
\centering
\caption{\textbf{Agent-direct control on RoboTwin~2.0 and RoboDojo.} Configurations passed on the first attempt and within three attempts, over all 204 trials. RoboDojo has 42 tasks, and each of its 12 generalization tasks has two layouts.}
\label{tab:supp_cl_combined}
\vspace{2pt}
\small
\setlength{\tabcolsep}{4pt}
\renewcommand{\arraystretch}{1.05}
\sbox{\clboxa}{\begin{tabular}[t]{@{}l c c r@{/}l c r@{/}l c@{}}
\toprule
 & & & \multicolumn{3}{c}{\textbf{First}} & \multicolumn{3}{c}{\textbf{Within 3}} \\
\cmidrule(lr){4-6}\cmidrule(l){7-9}
\textbf{Benchmark} & \textbf{Configs} & \textbf{Trials} & \multicolumn{2}{c}{Passed} & \% & \multicolumn{2}{c}{Passed} & \% \\
\midrule
RoboTwin 2.0 & 50 & 84 & 31 & 50 & 62.0 & 36 & 50 & 72.0 \\
\addlinespace[2pt]
RoboDojo & 54 & 120 & 20 & 54 & 37.0 & 26 & 54 & 48.1 \\
\quad Generalization & 24 & 51 & 10 & 24 & 41.7 & 13 & 24 & 54.2 \\
\quad Memory & 6 & 12 & 3 & 6 & 50.0 & 3 & 6 & 50.0 \\
\quad Open-vocabulary & 8 & 14 & 5 & 8 & 62.5 & 5 & 8 & 62.5 \\
\quad Long-horizon & 8 & 20 & 2 & 8 & 25.0 & 3 & 8 & 37.5 \\
\quad Precision & 8 & 23 & 0 & 8 & 0.0 & 2 & 8 & 25.0 \\
\bottomrule
\end{tabular}}%
\makebox[\linewidth][c]{\usebox{\clboxa}}
\end{table}}
\def\cltabletime{%
\begin{table}[t]
\centering
\caption{\textbf{Time and API cost of agent-direct control.} All quantities are medians over the trials of each benchmark, except the time in language-model inference, which is a ratio of sums, and the API cost, which is a mean, both over the trials with an API record.}
\label{tab:supp_cl_time}
\vspace{2pt}
\small
\setlength{\tabcolsep}{8pt}
\renewcommand{\arraystretch}{1.05}
\sbox{\clboxc}{\begin{tabular}[t]{@{}l c c@{}}
\toprule
\textbf{Quantity} & \textbf{RoboTwin 2.0} & \textbf{RoboDojo} \\
\midrule
Time per trial (min) & 18.2 & 32.8 \\
Commands per trial & 38 & 70 \\
Language-model calls per trial & 65.5 & 86.0 \\
Time per command (s) & 28.0 & 25.8 \\
Time in language-model inference (\%) & 93.0 & 86.5 \\
Robot motion time per trial (s) & 26.4 & 49.6 \\
API cost per trial (\$) & 3.27 & 4.35 \\
\bottomrule
\end{tabular}}%
\usebox{\clboxc}
\end{table}}

%% file: sections/supp/box_cl_prompt.tex
\providecommand{\pnote}[1]{\colorbox{black!7}{\rmfamily\color{black!60}#1}}
\providecommand{\phead}[1]{\medskip\noindent\pnote{\textbf{\small #1}}\par\nopagebreak\smallskip}
\providecommand{\pph}[1]{{\rmfamily\itshape\color{black!55}$\langle$#1$\rangle$}}
\providecommand{\pus}{\textunderscore\allowbreak}
\providecommand{\pnotepar}[1]{\par\noindent\colorbox{black!7}{\parbox{\dimexpr\linewidth-2\fboxsep\relax}{\rmfamily\footnotesize\color{black!60}#1}}\par}
\captionof{table}{\textbf{The prompt of the agent in agent-direct control.} The prompt contains only general rules and no task-specific procedure, and the task instruction is inserted at the placeholder. The name of the client script is replaced by a placeholder in angle brackets, shaded headings are the section titles of the prompt itself, shaded text is our annotation, and Markdown formatting marks are removed.}
\label{fig:supp_cl_prompt}
\nopagebreak
\begin{tcolorbox}[enhanced, breakable, colback=white, colframe=black!55, boxrule=0.5pt, arc=1pt,
 left=6pt, right=6pt, top=4pt, bottom=4pt, before skip=8pt, after skip=8pt,
 title after break={Tab.~\ref{fig:supp_cl_prompt} (continued)}, colbacktitle=black!7, coltitle=black!70,
 fonttitle=\footnotesize\rmfamily\bfseries, toptitle=1pt, bottomtitle=1pt]
\footnotesize\ttfamily\raggedright\frenchspacing\setlength{\parskip}{3pt}\widowpenalty=10000\clubpenalty=10000
\pnotepar{Opening paragraph of the prompt, without its title line.}

You are an autonomous robot operator. A robot with two arms, each with a parallel-jaw gripper and a wrist RGB camera, stands in front of a work area, and fixed RGB cameras also see the workspace. It is live and waiting for your commands. The robot is simulated, but you must treat it exactly like a real one: there is no reset, no undo, and no access to the simulator's internal state {\rmfamily\textemdash} only the cameras and the robot's own proprioception. Nothing about the scene is given to you in advance.

\phead{Task}

\pph{task instruction}

The sentence above is the whole goal. Exactly how you achieve the stated end state {\rmfamily\textemdash} including which arm (or both) you use {\rmfamily\textemdash} is your choice; that the end state holds, is not. If the sentence asks you to hold, lift, pick up or carry something, the end state is holding it (keep the grip, do not put it down); otherwise finish with both grippers open and both arms back in their start posture (home with "arm":"both").

\pnotepar{The Interface section, which names the client script and the file that documents the 11 commands, is omitted.}

\phead{Rules}

1. Modify files ONLY inside this directory. Put every script, note and extracted frame you create under scratch/.\\
2. Interact with the robot ONLY via \pph{client script} and the files it returns (frames/\ldots). Experiment-integrity rule: do NOT read the robot software's source code or any file outside this directory, and do not connect to any other machine {\rmfamily\textemdash} everything you legitimately need is in here. Violating this voids the experiment.\\
3. Budget: at most 300 counted commands and 45 minutes of wall clock. Look (frames) before and after every consequential action and verify visually; the gripper's returned opening after close tells you whether you are actually holding something.\\
4. If an approach fails repeatedly, change strategy rather than repeating the same command. There is NO way to reset the scene. If an object slips, drops, tips or ends up somewhere unexpected, deal with the scene AS IT IS: go back to an observation posture, re-perceive, and continue from the new state. Never push the grasp point or the fingers through a supporting surface: surface heights are not given, determine them yourself from the images and from contact.\\
5. Operation log (mandatory, part of the experiment record): maintain scratch/OPERATION\pus{}LOG.md like a lab notebook. Before each motion command (move\pus{}ee, move\pus{}delta, move\pus{}joints, home, gripper) append one numbered entry with: the arm, the frames used (paths), the scene measurements derived from them (object/\allowbreak{}target positions and sizes in metres), the commanded target, and the expected outcome in measurable terms. After the result returns, append the actual outcome (returned pose, error, gripper opening, what the next frames show) and, if it differed from the expectation, the correction you apply next. Plain prose, specific numbers.

\phead{Deliverable}

When the task is done {\rmfamily\textemdash} or you conclude it cannot be done {\rmfamily\textemdash} write scratch/RESULT.md: what you understood the goal to be, your plan, what actually happened (with key frames/ paths), and your own honest judgement of success. Success means: the stated end state holds (still held if the task is to hold/\allowbreak{}lift/\allowbreak{}pick up; otherwise released, grippers open, both arms home), confirmed in a final image from a fixed camera. Then stop.
\end{tcolorbox}

%% file: main.bbl
\begin{thebibliography}{52}
\providecommand{\natexlab}[1]{#1}
\providecommand{\url}[1]{\texttt{#1}}
\expandafter\ifx\csname urlstyle\endcsname\relax
  \providecommand{\doi}[1]{doi: #1}\else
  \providecommand{\doi}{doi: \begingroup \urlstyle{rm}\Url}\fi

\bibitem[Ahn et~al.(2022)Ahn, Brohan, Brown, Chebotar, Cortes, David, Finn, Fu,
  Gopalakrishnan, Hausman, Herzog, Ho, Hsu, Ibarz, Ichter, Irpan, Jang, Ruano,
  Jeffrey, Jesmonth, Joshi, Julian, Kalashnikov, Kuang, Lee, Levine, Lu, Luu,
  Parada, Pastor, Quiambao, Rao, Rettinghouse, Reyes, Sermanet, Sievers, Tan,
  Toshev, Vanhoucke, Xia, Xiao, Xu, Xu, Yan, and Zeng]{saycan2022}
Michael Ahn, Anthony Brohan, Noah Brown, Yevgen Chebotar, Omar Cortes, Byron
  David, Chelsea Finn, Chuyuan Fu, Keerthana Gopalakrishnan, Karol Hausman,
  Alex Herzog, Daniel Ho, Jasmine Hsu, Julian Ibarz, Brian Ichter, Alex Irpan,
  Eric Jang, Rosario~Jauregui Ruano, Kyle Jeffrey, Sally Jesmonth, Nikhil~J
  Joshi, Ryan Julian, Dmitry Kalashnikov, Yuheng Kuang, Kuang-Huei Lee, Sergey
  Levine, Yao Lu, Linda Luu, Carolina Parada, Peter Pastor, Jornell Quiambao,
  Kanishka Rao, Jarek Rettinghouse, Diego Reyes, Pierre Sermanet, Nicolas
  Sievers, Clayton Tan, Alexander Toshev, Vincent Vanhoucke, Fei Xia, Ted Xiao,
  Peng Xu, Sichun Xu, Mengyuan Yan, and Andy Zeng.
\newblock {Do As I Can, Not As I Say: Grounding Language in Robotic
  Affordances}.
\newblock In \emph{Conference on Robot Learning (CoRL)}, 2022.

\bibitem[Black et~al.(2025{\natexlab{a}})Black, Brown, Darpinian, Dhabalia,
  Driess, Esmail, Equi, Finn, Fusai, Galliker, Ghosh, Groom, Hausman, Ichter,
  Jakubczak, Jones, Ke, LeBlanc, Levine, Li-Bell, Mothukuri, Nair, Pertsch,
  Ren, Shi, Smith, Springenberg, Stachowicz, Tanner, Vuong, Walke, Walling,
  Wang, Yu, and Zhilinsky]{pi05_2025}
Kevin Black, Noah Brown, James Darpinian, Karan Dhabalia, Danny Driess, Adnan
  Esmail, Michael Equi, Chelsea Finn, Niccolo Fusai, Manuel~Y. Galliker, Dibya
  Ghosh, Lachy Groom, Karol Hausman, Brian Ichter, Szymon Jakubczak, Tim Jones,
  Liyiming Ke, Devin LeBlanc, Sergey Levine, Adrian Li-Bell, Mohith Mothukuri,
  Suraj Nair, Karl Pertsch, Allen~Z. Ren, Lucy~Xiaoyang Shi, Laura Smith,
  Jost~Tobias Springenberg, Kyle Stachowicz, James Tanner, Quan Vuong, Homer
  Walke, Anna Walling, Haohuan Wang, Lili Yu, and Ury Zhilinsky.
\newblock {$\pi_{0.5}$: a Vision-Language-Action Model with Open-World
  Generalization}.
\newblock In \emph{Conference on Robot Learning (CoRL)}, 2025{\natexlab{a}}.

\bibitem[Black et~al.(2025{\natexlab{b}})Black, Brown, Driess, Esmail, Equi,
  Finn, Fusai, Groom, Hausman, Ichter, Jakubczak, Jones, Ke, Levine, Li-Bell,
  Mothukuri, Nair, Pertsch, Shi, Tanner, Vuong, Walling, Wang, and
  Zhilinsky]{pi0_2024}
Kevin Black, Noah Brown, Danny Driess, Adnan Esmail, Michael Equi, Chelsea
  Finn, Niccolo Fusai, Lachy Groom, Karol Hausman, Brian Ichter, Szymon
  Jakubczak, Tim Jones, Liyiming Ke, Sergey Levine, Adrian Li-Bell, Mohith
  Mothukuri, Suraj Nair, Karl Pertsch, Lucy~Xiaoyang Shi, James Tanner, Quan
  Vuong, Anna Walling, Haohuan Wang, and Ury Zhilinsky.
\newblock {$\pi_0$: A Vision-Language-Action Flow Model for General Robot
  Control}.
\newblock In \emph{Robotics: Science and Systems (RSS)}, 2025{\natexlab{b}}.

\bibitem[Brohan et~al.(2023)Brohan, Brown, Carbajal, Chebotar, Chen,
  Choromanski, Ding, Driess, Dubey, Finn, Florence, Fu, Arenas, Gopalakrishnan,
  Han, Hausman, Herzog, Hsu, Ichter, Irpan, Joshi, Julian, Kalashnikov, Kuang,
  Leal, Lee, Lee, Levine, Lu, Michalewski, Mordatch, Pertsch, Rao, Reymann,
  Ryoo, Salazar, Sanketi, Sermanet, Singh, Singh, Soricut, Tran, Vanhoucke,
  Vuong, Wahid, Welker, Wohlhart, Wu, Xia, Xiao, Xu, Xu, Yu, and
  Zitkovich]{rt2_2023}
Anthony Brohan, Noah Brown, Justice Carbajal, Yevgen Chebotar, Xi Chen,
  Krzysztof Choromanski, Tianli Ding, Danny Driess, Avinava Dubey, Chelsea
  Finn, Pete Florence, Chuyuan Fu, Montse~Gonzalez Arenas, Keerthana
  Gopalakrishnan, Kehang Han, Karol Hausman, Alexander Herzog, Jasmine Hsu,
  Brian Ichter, Alex Irpan, Nikhil Joshi, Ryan Julian, Dmitry Kalashnikov,
  Yuheng Kuang, Isabel Leal, Lisa Lee, Tsang-Wei~Edward Lee, Sergey Levine, Yao
  Lu, Henryk Michalewski, Igor Mordatch, Karl Pertsch, Kanishka Rao, Krista
  Reymann, Michael Ryoo, Grecia Salazar, Pannag Sanketi, Pierre Sermanet,
  Jaspiar Singh, Anikait Singh, Radu Soricut, Huong Tran, Vincent Vanhoucke,
  Quan Vuong, Ayzaan Wahid, Stefan Welker, Paul Wohlhart, Jialin Wu, Fei Xia,
  Ted Xiao, Peng Xu, Sichun Xu, Tianhe Yu, and Brianna Zitkovich.
\newblock {RT-2: Vision-Language-Action Models Transfer Web Knowledge to
  Robotic Control}.
\newblock In \emph{Conference on Robot Learning (CoRL)}, 2023.

\bibitem[Carion et~al.(2025)Carion, Gustafson, Hu, Debnath, Hu, Suris, Ryali,
  Alwala, Khedr, Huang, Lei, Ma, Guo, Kalla, Marks, Greer, Wang, Sun,
  R{\"a}dle, Afouras, Mavroudi, Xu, Wu, Zhou, Momeni, Hazra, Ding, Vaze,
  Porcher, Li, Li, Kamath, Cheng, Doll{\'a}r, Ravi, Saenko, Zhang, and
  Feichtenhofer]{carion2025sam3}
Nicolas Carion, Laura Gustafson, Yuan-Ting Hu, Shoubhik Debnath, Ronghang Hu,
  Didac Suris, Chaitanya Ryali, Kalyan~Vasudev Alwala, Haitham Khedr, Andrew
  Huang, Jie Lei, Tengyu Ma, Baishan Guo, Arpit Kalla, Markus Marks, Joseph
  Greer, Meng Wang, Peize Sun, Roman R{\"a}dle, Triantafyllos Afouras,
  Effrosyni Mavroudi, Katherine Xu, Tsung-Han Wu, Yu Zhou, Liliane Momeni,
  Rishi Hazra, Shuangrui Ding, Sagar Vaze, Francois Porcher, Feng Li, Siyuan
  Li, Aishwarya Kamath, Ho~Kei Cheng, Piotr Doll{\'a}r, Nikhila Ravi, Kate
  Saenko, Pengchuan Zhang, and Christoph Feichtenhofer.
\newblock {SAM 3: Segment Anything with Concepts}.
\newblock \emph{arXiv preprint arXiv:2511.16719}, 2025.

\bibitem[Chen et~al.(2025)Chen, Lessing, Liu, and Finn]{chen2025curating}
Annie~S. Chen, Alec~M. Lessing, Yuejiang Liu, and Chelsea Finn.
\newblock {Curating Demonstrations using Online Experience}.
\newblock In \emph{Robotics: Science and Systems (RSS)}, 2025.

\bibitem[Chen et~al.(2026{\natexlab{a}})Chen, Hadfield, Zook, Uy, Song,
  Coumans, Yang, Ladhak, Qu, Birchfield, Tremblay, and Blukis]{chen2026volo}
Siyi Chen, Hugo Hadfield, Alex Zook, Mikaela~Angelina Uy, Chan~Hee Song, Erwin
  Coumans, Xuning Yang, Faisal Ladhak, Qing Qu, Stan Birchfield, Jonathan
  Tremblay, and Valts Blukis.
\newblock {VoLo: A Physical Orchestrator for Open-Vocabulary Long-Horizon
  Manipulation}.
\newblock \emph{arXiv preprint arXiv:2606.07723}, 2026{\natexlab{a}}.

\bibitem[Chen et~al.(2026{\natexlab{b}})Chen, Chen, Li, Tang, Su, Lu, Wan,
  Chen, Liu, Yan, et~al.]{chen2026robodojo}
Tianxing Chen, Yue Chen, Zixuan Li, Junyuan Tang, Kailun Su, Haoran Lu, Weijie
  Wan, Baijun Chen, Songling Liu, Haowen Yan, et~al.
\newblock {RoboDojo}: A unified sim-and-real benchmark for comprehensive
  evaluation of generalist robot manipulation policies.
\newblock \emph{arXiv preprint arXiv:2607.04434}, 2026{\natexlab{b}}.

\bibitem[Chen et~al.(2026{\natexlab{c}})Chen, Chen, Chen, Cai, Liu, Li, Liang,
  Lin, Ge, Gu, Deng, Guo, Nian, Xie, Chen, Su, Xu, Liu, Hu, Gao, Wang, Liang,
  Qin, Yang, Luo, and Mu]{robotwin2_2025}
Tianxing Chen, Zanxin Chen, Baijun Chen, Zijian Cai, Yibin Liu, Zixuan Li,
  Qiwei Liang, Xianliang Lin, Yiheng Ge, Zhenyu Gu, Weiliang Deng, Yubin Guo,
  Tian Nian, Xuanbing Xie, Qiangyu Chen, Kailun Su, Tianling Xu, Guodong Liu,
  Mengkang Hu, Huan-ang Gao, Kaixuan Wang, Zhixuan Liang, Yusen Qin, Xiaokang
  Yang, Ping Luo, and Yao Mu.
\newblock {RoboTwin 2.0: A Scalable Data Generator and Benchmark with Strong
  Domain Randomization for Robust Bimanual Robotic Manipulation}.
\newblock In \emph{International Conference on Machine Learning (ICML)},
  2026{\natexlab{c}}.

\bibitem[Galanti et~al.(2026)Galanti, Shah, and Dao]{galanti2026gap}
Liane Galanti, Dhruv Shah, and Tri Dao.
\newblock {Addressing the Orchestration Gap in Generalist Robots via Physical
  Agency}.
\newblock \emph{arXiv preprint arXiv:2607.21725}, 2026.

\bibitem[Gu et~al.(2026)Gu, Yang, Zou, and Xu]{primskills2026}
Zhihao Gu, Ming Yang, Difan Zou, and Dong Xu.
\newblock {Learning Diffusion Policy from Primitive Skills for Robot
  Manipulation}.
\newblock In \emph{AAAI Conference on Artificial Intelligence (AAAI)}, 2026.

\bibitem[Guo et~al.(2026)Guo, Yan, Zhong, Zhang, Yang, Lu, Chen, Wang, Deng,
  Yang, Yuan, Liu, Ban, Zhu, Liu, Feng, Li, and Chen]{guo2026vpp2}
Yanjiang Guo, Haodong Yan, Zhide Zhong, Zhongru Zhang, Qingyuan Yang, Qingzhou
  Lu, Xiaoyu Chen, Yen-Jen Wang, Shuying Deng, Chenghan Yang, Puzhen Yuan,
  Chenxin Liu, Tun Ban, Xiang Zhu, Yichen Liu, Kun Feng, Haoang Li, and Jianyu
  Chen.
\newblock {Video Prediction Policy 2: Predict Better, Act Better}.
\newblock \emph{arXiv preprint arXiv:2610.10270}, 2026.

\bibitem[Hao et~al.(2026)Hao, Syed, Schneider, and
  Ichnowski]{skillretrieval2026}
Haoran Hao, Shahram~Najam Syed, Jeff Schneider, and Jeffrey Ichnowski.
\newblock {Hierarchical Skill Retrieval for Data-Efficient Adaptation of
  Vision-Language-Action Models}.
\newblock \emph{arXiv preprint arXiv:2608.24042}, 2026.

\bibitem[Hu et~al.(2026)Hu, Shridhar, Lu, Shah, Chiang, Tan, and
  Xie]{hu2026orchestrating}
Jiaheng Hu, Mohit Shridhar, Caden Lu, Dhruv Shah, Hao-Tien~Lewis Chiang, Jie
  Tan, and Annie Xie.
\newblock {What Matters in Orchestrating Robot Policies: A Systematic Study of
  Hierarchical VLA Agents}.
\newblock \emph{arXiv preprint arXiv:2606.10267}, 2026.

\bibitem[Huang et~al.(2026)Huang, Li, Hu, Qi, Xiao, Coates, Cao, Zhang, and
  Zhang]{huang2026roboharness}
Jinbang Huang, Zhiyuan Li, Yuanzhao Hu, Ran Qi, Yixin Xiao, Mark Coates,
  Tongtong Cao, Zhanguang Zhang, and Yingxue Zhang.
\newblock {RoboHarness: Memory-Driven Orchestration of Heterogeneous Robot
  Policies for Long-Horizon Planning}.
\newblock \emph{arXiv preprint arXiv:2607.18060}, 2026.

\bibitem[Huang et~al.(2022)Huang, Xia, Xiao, Chan, Liang, Florence, Zeng,
  Tompson, Mordatch, Chebotar, Sermanet, Brown, Jackson, Luu, Levine, Hausman,
  and Ichter]{innermonologue2022}
Wenlong Huang, Fei Xia, Ted Xiao, Harris Chan, Jacky Liang, Pete Florence, Andy
  Zeng, Jonathan Tompson, Igor Mordatch, Yevgen Chebotar, Pierre Sermanet, Noah
  Brown, Tomas Jackson, Linda Luu, Sergey Levine, Karol Hausman, and Brian
  Ichter.
\newblock {Inner Monologue: Embodied Reasoning through Planning with Language
  Models}.
\newblock In \emph{Conference on Robot Learning (CoRL)}, 2022.

\bibitem[Huang et~al.(2023)Huang, Wang, Zhang, Li, Wu, and
  Fei-Fei]{voxposer2023}
Wenlong Huang, Chen Wang, Ruohan Zhang, Yunzhu Li, Jiajun Wu, and Li Fei-Fei.
\newblock {VoxPoser: Composable 3D Value Maps for Robotic Manipulation with
  Language Models}.
\newblock In \emph{Conference on Robot Learning (CoRL)}, 2023.

\bibitem[Jia et~al.(2026)Jia, Lin, Zhang, Zhang, Liu, and Jiang]{jia2026agp}
Mengzhao Jia, Yang Lin, Xixin Zhang, Zhihan Zhang, Xiaobai Liu, and Meng Jiang.
\newblock {Agent as Policy for Robotic Manipulation}.
\newblock \emph{arXiv preprint arXiv:2609.12541}, 2026.

\bibitem[Jin and Zhang(2026)]{agenticvla2026}
Ruofan Jin and Zaixi Zhang.
\newblock {Agentic-VLA: Efficient Online Adaptation for Vision-Language-Action
  Models}.
\newblock \emph{arXiv preprint arXiv:2605.22896}, 2026.

\bibitem[Kim et~al.(2024)Kim, Pertsch, Karamcheti, Xiao, Balakrishna, Nair,
  Rafailov, Foster, Sanketi, Vuong, Kollar, Burchfiel, Tedrake, Sadigh, Levine,
  Liang, and Finn]{openvla2024}
Moo~Jin Kim, Karl Pertsch, Siddharth Karamcheti, Ted Xiao, Ashwin Balakrishna,
  Suraj Nair, Rafael Rafailov, Ethan Foster, Pannag Sanketi, Quan Vuong, Thomas
  Kollar, Benjamin Burchfiel, Russ Tedrake, Dorsa Sadigh, Sergey Levine, Percy
  Liang, and Chelsea Finn.
\newblock {OpenVLA: An Open-Source Vision-Language-Action Model}.
\newblock In \emph{Conference on Robot Learning (CoRL)}, 2024.

\bibitem[Kim et~al.(2027)Kim, Bhat, Krishnamurthy, Karri, and
  Khorrami]{claro2026}
Sungsu Kim, Vineet Bhat, Prashanth Krishnamurthy, Ramesh Karri, and Farshad
  Khorrami.
\newblock {CLARO: Clutter Aware Robot Orchestrator for Open-Vocabulary Grasping
  using VLM Agents}.
\newblock In \emph{IEEE/CVF Winter Conference on Applications of Computer
  Vision (WACV)}, 2027.

\bibitem[Konidaris and Barto(2009)]{konidaris2009skillchaining}
George Konidaris and Andrew~G. Barto.
\newblock {Skill Discovery in Continuous Reinforcement Learning Domains using
  Skill Chaining}.
\newblock In \emph{Advances in Neural Information Processing Systems}, pages
  1015--1023, 2009.

\bibitem[Levine et~al.(2020)Levine, Kumar, Tucker, and Fu]{levine2020offline}
Sergey Levine, Aviral Kumar, George Tucker, and Justin Fu.
\newblock {Offline Reinforcement Learning: Tutorial, Review, and Perspectives
  on Open Problems}.
\newblock \emph{arXiv preprint arXiv:2005.01643}, 2020.

\bibitem[Li et~al.(2026{\natexlab{a}})Li, Mishra, Liu, Tao, Kumar, Gray,
  Wickramarachchi, Francis, Scherer, and Silver]{li2026recover}
Bowen Li, Mayank Mishra, Y.~Isabel Liu, Stone Tao, Nishanth Kumar, Alexander~G.
  Gray, Ruwan Wickramarachchi, Jonathan Francis, Sebastian Scherer, and Tom
  Silver.
\newblock {Recover, Discover, Plan: Learning Skills and Concepts from Robot
  Failures}.
\newblock In \emph{Conference on Robot Learning (CoRL)}, 2026{\natexlab{a}}.

\bibitem[Li et~al.(2025)Li, Peng, Li, Qiao, Zheng, Sun, Qin, Li, Luan, Wu,
  Zhan, Sun, Xu, Li, Shen, and He]{atomiclib2025}
Dongjiang Li, Bo Peng, Chang Li, Ning Qiao, Qi Zheng, Lei Sun, Yusen Qin,
  Bangguo Li, Yifeng Luan, Bo Wu, Yibing Zhan, Mingang Sun, Tong Xu, Lusong Li,
  Hui Shen, and Xiaodong He.
\newblock {An Atomic Skill Library Construction Method for Data-Efficient
  Embodied Manipulation}.
\newblock \emph{arXiv preprint arXiv:2501.15068}, 2025.

\bibitem[Li et~al.(2026{\natexlab{b}})Li, Zhou, Zhu, Chen, Wang, Wang, Hu, Yu,
  Jiang, Su, Ma, He, Shen, Yang, Ren, Yao, Wang, and Mu]{roboclaw2026}
Ruiying Li, Yunlang Zhou, YuYao Zhu, Kylin Chen, Jingyuan Wang, Sukai Wang,
  Kongtao Hu, Minhui Yu, Bowen Jiang, Zhan Su, Jiayao Ma, Xin He, Yongjian
  Shen, Yang Yang, Guanghui Ren, Maoqing Yao, Wenhao Wang, and Yao Mu.
\newblock {RoboClaw: An Agentic Framework for Scalable Long-Horizon Robotic
  Tasks}.
\newblock \emph{arXiv preprint arXiv:2603.11558}, 2026{\natexlab{b}}.

\bibitem[Liang et~al.(2023)Liang, Huang, Xia, Xu, Hausman, Ichter, Florence,
  and Zeng]{codeaspolicies2023}
Jacky Liang, Wenlong Huang, Fei Xia, Peng Xu, Karol Hausman, Brian Ichter, Pete
  Florence, and Andy Zeng.
\newblock {Code as Policies: Language Model Programs for Embodied Control}.
\newblock In \emph{IEEE International Conference on Robotics and Automation
  (ICRA)}, 2023.

\bibitem[Lu et~al.(2026)Lu, Wu, Kou, Lin, Fu, Xiao, Mandlekar, Xu, Shi,
  Goldberg, Chen, Chowdhury, Kautz, Zhu, Fan, and Wang]{aspire2026}
Runyu Lu, Yubo Wu, Ethan Kou, Haotian Lin, Letian Fu, Wenli Xiao, Ajay
  Mandlekar, Yinzhen Xu, Guanya Shi, Ken Goldberg, Ang Chen, Mosharaf
  Chowdhury, Jan Kautz, Yuke Zhu, Linxi~``Jim'' Fan, and Guanzhi Wang.
\newblock {ASPIRE: Agentic /Skills Discovery for Robotics}.
\newblock \emph{arXiv preprint arXiv:2607.00272}, 2026.

\bibitem[Mao et~al.(2024)Mao, Zhong, Jiang, Fang, Zhang, Lan, Li, Jia, Wang,
  Fan, and Yoshie]{robomatrix2024}
Weixin Mao, Weiheng Zhong, Zhou Jiang, Dong Fang, Zhongyue Zhang, Zihan Lan,
  Haosheng Li, Fan Jia, Tiancai Wang, Haoqiang Fan, and Osamu Yoshie.
\newblock {RoboMatrix: A Skill-centric Hierarchical Framework for Scalable
  Robot Task Planning and Execution in Open-World}.
\newblock \emph{arXiv preprint arXiv:2412.00171}, 2024.

\bibitem[{Octo Model Team} et~al.(2024){Octo Model Team}, Ghosh, Walke,
  Pertsch, Black, Mees, Dasari, Hejna, Kreiman, Xu, Luo, Tan, Chen, Sanketi,
  Vuong, Xiao, Sadigh, Finn, and Levine]{octo2024}
{Octo Model Team}, Dibya Ghosh, Homer Walke, Karl Pertsch, Kevin Black, Oier
  Mees, Sudeep Dasari, Joey Hejna, Tobias Kreiman, Charles Xu, Jianlan Luo,
  You~Liang Tan, Lawrence~Yunliang Chen, Pannag Sanketi, Quan Vuong, Ted Xiao,
  Dorsa Sadigh, Chelsea Finn, and Sergey Levine.
\newblock {Octo: An Open-Source Generalist Robot Policy}.
\newblock In \emph{Robotics: Science and Systems (RSS)}, 2024.

\bibitem[Ross et~al.(2011)Ross, Gordon, and Bagnell]{ross2011dagger}
St{\'e}phane Ross, Geoffrey~J. Gordon, and J.~Andrew Bagnell.
\newblock {A Reduction of Imitation Learning and Structured Prediction to
  No-Regret Online Learning}.
\newblock In \emph{Proceedings of the Fourteenth International Conference on
  Artificial Intelligence and Statistics (AISTATS)}, pages 627--635, 2011.

\bibitem[Rui et~al.(2026)Rui, Zuo, Wang, Jia, Ma, Zhou, and
  Li]{semantichandoff2026}
Ke Rui, Yushen Zuo, Jiawei Wang, Haoran Jia, Jinming Ma, Weitao Zhou, and
  Minglei Li.
\newblock {Diagnosing Semantic Handoff Failures in Agent-Orchestrated
  Vision-Language-Action Skill Composition}.
\newblock \emph{arXiv preprint arXiv:2607.06256}, 2026.

\bibitem[Santos et~al.(2026)Santos, Khadraoui, Farhat, Yous, Teffahi, and
  Hacid]{alrm2026}
Vitor Gaboardi~dos Santos, Ibrahim Khadraoui, Ibrahim Farhat, Hamza Yous, Samy
  Teffahi, and Hakim Hacid.
\newblock {ALRM: Agentic LLM for Robotic Manipulation}.
\newblock \emph{arXiv preprint arXiv:2601.19510}, 2026.

\bibitem[SenseNova et~al.(2026)SenseNova, Team, Cai, Cao, Chen, Chua, Dai,
  Deng, Deng, Diao, Ding, Fan, Fu, Gong, Gu, He, Hou, Li, Li, Li, Li, Liao,
  Lin, Lin, Liu, Liu, Liu, Liu, Liu, Loy, Lu, Luo, Ma, Pan, Pan, Qian, Shang,
  Shao, Su, Tao, Tao, Wang, Wang, Wang, Wang, Wen, Wu, Wu, Wu, Wu, Wu, Xiao,
  Xu, Yamashita, Yang, Yang, Yin, Yin, Zhao, Zhong, Zhuo, and
  Zuo]{sensenova2026astra}
SenseNova, Joint~Research Team, Zhongang Cai, Ziang Cao, Tailai Chen, Han~Chong
  Chua, Linjun Dai, Hanming Deng, Hokin Deng, Haiwen Diao, Chenjing Ding,
  Weichen Fan, Xiao Fu, Zerui Gong, Chenyang Gu, Zexin He, Zhi Hou, Bo Li,
  Qinyue Li, Quanzhou Li, Xiaowan Li, Zhichao Liao, Dahua Lin, Zilong Lin, Bo
  Liu, Chunxiao Liu, Pengfei Liu, Tianqi Liu, Ziwei Liu, Chen~Change Loy, Lewei
  Lu, Jiapeng Luo, Yuexiao Ma, Liang Pan, Xingang Pan, Oscar Qian, Ziyi Shang,
  Yuyao Shao, Dawei Su, Dacheng Tao, Qingyi Tao, Quan Wang, Ruisi Wang,
  Xiaogang Wang, Ziyi Wang, Beichen Wen, Donghao Wu, Jiaxu Wu, Silei Wu,
  Xiaoming Wu, Zhonghua Wu, Hongcan Xiao, Weiye Xu, Atsushi Yamashita, Lei
  Yang, Zhe Yang, Baiqiao Yin, Wanqi Yin, Hanfei Zhao, Tianbo Zhong, Long Zhuo,
  and Jing Zuo.
\newblock {Has GPT-6-Astra Achieved Embodied Intelligence?}
\newblock \emph{arXiv preprint arXiv:2610.22471}, 2026.

\bibitem[Shi et~al.(2026)Shi, Yang, Chao, Wan, Shao, Lei, Qian, Le, Chaudhari,
  Daniilidis, Wen, and Jayaraman]{maestro2025}
Junyao Shi, Rujia Yang, Kaitian Chao, Selina~Bingqing Wan, Yifei Shao, Jiahui
  Lei, Jianing Qian, Long Le, Pratik Chaudhari, Kostas Daniilidis, Chuan Wen,
  and Dinesh Jayaraman.
\newblock {MAESTRO: Orchestrating Robotics Modules with Coding Agents for
  Zero-Shot Generalist Robots}.
\newblock In \emph{Conference on Robot Learning (CoRL)}, 2026.

\bibitem[Shinn et~al.(2023)Shinn, Cassano, Gopinath, Narasimhan, and
  Yao]{reflexion2023}
Noah Shinn, Federico Cassano, Ashwin Gopinath, Karthik Narasimhan, and Shunyu
  Yao.
\newblock {Reflexion: Language Agents with Verbal Reinforcement Learning}.
\newblock In \emph{Advances in Neural Information Processing Systems
  (NeurIPS)}, 2023.

\bibitem[Sukhija et~al.(2026)Sukhija, Groth, Shridhar, Hertweck, Bloesch,
  Wulfmeier, Abdolmaleki, and Riedmiller]{eximo2026}
Bhavya Sukhija, Oliver Groth, Mohit Shridhar, Tim Hertweck, Michael Bloesch,
  Markus Wulfmeier, Abbas Abdolmaleki, and Martin Riedmiller.
\newblock {EXIMO: VLM Guided Exploration of VLA Policies}.
\newblock \emph{arXiv preprint arXiv:2608.19891}, 2026.

\bibitem[Sutton et~al.(1999)Sutton, Precup, and Singh]{sutton1999options}
Richard~S. Sutton, Doina Precup, and Satinder Singh.
\newblock {Between MDPs and Semi-MDPs: A Framework for Temporal Abstraction in
  Reinforcement Learning}.
\newblock \emph{Artificial Intelligence}, 112\penalty0 (1--2):\penalty0
  181--211, 1999.

\bibitem[Wang et~al.(2024)Wang, Xie, Jiang, Mandlekar, Xiao, Zhu, Fan, and
  Anandkumar]{voyager2023}
Guanzhi Wang, Yuqi Xie, Yunfan Jiang, Ajay Mandlekar, Chaowei Xiao, Yuke Zhu,
  Linxi Fan, and Anima Anandkumar.
\newblock {Voyager: An Open-Ended Embodied Agent with Large Language Models}.
\newblock \emph{Transactions on Machine Learning Research (TMLR)}, 2024.

\bibitem[Wang et~al.(2026)Wang, Osterberg, Tian, Shorinwa, Bowkett, Pastor, Wu,
  and Schwager]{insight2026}
Maggie Wang, Lars Osterberg, Stephen Tian, Ola Shorinwa, Joseph Bowkett, Daniel
  Pastor, Jiajun Wu, and Mac Schwager.
\newblock {InSight: Self-Guided Skill Acquisition via Steerable VLAs}.
\newblock \emph{arXiv preprint arXiv:2606.24884}, 2026.

\bibitem[Wu et~al.(2026)Wu, Lu, Wang, Yang, Liu, Wang, Zhu, Sun, Wang, Ma, Ren,
  Zhang, Yu, Zhao, Zhou, Qiu, Xiong, Wang, Wang, Cheng, Li, Huang, Zhu, Shen,
  and Zheng]{lingbotvla2026}
Wei Wu, Fan Lu, Yunnan Wang, Shuai Yang, Shi Liu, Fangjing Wang, Qian Zhu, He
  Sun, Yong Wang, Shuailei Ma, Yiyu Ren, Kejia Zhang, Hui Yu, Jingmei Zhao,
  Shuai Zhou, Zhenqi Qiu, Houlong Xiong, Ziyu Wang, Zechen Wang, Ran Cheng,
  Yong-Lu Li, Yongtao Huang, Xing Zhu, Yujun Shen, and Kecheng Zheng.
\newblock {A Pragmatic VLA Foundation Model}.
\newblock \emph{arXiv preprint arXiv:2601.18692}, 2026.

\bibitem[Xiao et~al.(2026)Xiao, Xie, Zhang, Lin, Fu, Xue, Lu, Yang, Dai, Wang,
  Wu, Wang, Sastry, Goldberg, Fan, Zhu, and Shi]{enpire2026}
Wenli Xiao, Jia Xie, Tonghe Zhang, Haotian Lin, Letian~``Max'' Fu, Haoru Xue,
  Jalen Lu, Yi Yang, Cunxi Dai, Zi Wang, Jimmy Wu, Guanzhi Wang, S.~Shankar
  Sastry, Ken Goldberg, Linxi~``Jim'' Fan, Yuke Zhu, and Guanya Shi.
\newblock {ENPIRE: Agentic Robot Policy Self-Improvement in the Real World}.
\newblock In \emph{Conference on Robot Learning (CoRL)}, 2026.

\bibitem[Xu et~al.(2026)Xu, Shang, and Ferrara]{xu2026baton}
Bingxin Xu, Yuzhang Shang, and Emilio Ferrara.
\newblock {Don't Drop the BATON: Long-Horizon Robot Manipulation via Agentic
  Subtask Exploration and Transition-aware Memory}.
\newblock \emph{arXiv preprint arXiv:2608.16889}, 2026.

\bibitem[Yang et~al.(2026{\natexlab{a}})Yang, Xu, Zhang, Tan, Zhang, Gong, and
  Xie]{mastvla2026}
Jiachi Yang, Zhiyi Xu, Hongquan Zhang, Xin Tan, Zhizhong Zhang, Jingyu Gong,
  and Yuan Xie.
\newblock {MAST-VLA: Skill-Conditioned Adaptation with Trajectory-Affordance
  Priors for Multi-Skill Robot Manipulation}.
\newblock In \emph{Conference on Robot Learning (CoRL)}, 2026{\natexlab{a}}.

\bibitem[Yang et~al.(2026{\natexlab{b}})Yang, Khetan, and Zhang]{astranomy2026}
Jenai~Xuning Yang, Ashna Khetan, and Shun Zhang.
\newblock {Astranomy: An Empirical Study of GPT-6 Astra as a Native Robot
  Policy}.
\newblock Technical report,
  \url{https://research.nvidia.com/labs/srl/projects/astranomy/},
  2026{\natexlab{b}}.
\newblock Accessed: 2026-11-15.

\bibitem[Yang et~al.(2026{\natexlab{c}})Yang, Chen, Gao, and Zhang]{ora2026}
Yuanbo Yang, Siyi Chen, Jun Gao, and Hao Zhang.
\newblock {ORA: Open Robot-Use Agents}.
\newblock Hao AI Lab blog, \url{https://haoailab.com/blogs/ora/},
  2026{\natexlab{c}}.
\newblock Accessed: 2026-11-15.

\bibitem[Zhai et~al.(2026)Zhai, Huang, Wu, Zhao, Yu, Ren, Hao, and
  Soh]{skillvla2026}
Xuanran Zhai, Zekai Huang, Longyan Wu, Qianyou Zhao, Qiaojun Yu, Jieji Ren, Ce
  Hao, and Harold Soh.
\newblock {SkillVLA: Tackling Combinatorial Diversity in Dual-Arm Manipulation
  via Skill Reuse}.
\newblock \emph{arXiv preprint arXiv:2603.03836}, 2026.

\bibitem[Zhang et~al.(2026{\natexlab{a}})Zhang, Tang, Zhan, Chen, Chen, Han,
  Zhu, Xu, Xu, Wu, Lin, and Liang]{atomicvla2026}
Likui Zhang, Tao Tang, Zhihao Zhan, Xiuwei Chen, Zisheng Chen, Jianhua Han,
  Jiangtong Zhu, Pei Xu, Hang Xu, Hefeng Wu, Liang Lin, and Xiaodan Liang.
\newblock {AtomicVLA: Unlocking the Potential of Atomic Skill Learning in
  Robots}.
\newblock In \emph{IEEE/CVF Conference on Computer Vision and Pattern
  Recognition (CVPR)}, 2026{\natexlab{a}}.

\bibitem[Zhang et~al.(2026{\natexlab{b}})Zhang, Wang, Ouyang, Huang, Li, Su,
  Jin, Chai, Liang, Dou, Chen, and Chen]{astra2026robodojo}
Wenbo Zhang, Kaixuan Wang, Yutao Ouyang, Xiaoyu Huang, Liyang Li, Kailun Su,
  Weiyang Jin, Wenhao Chai, Haotian Liang, Zhiyang Dou, Yue Chen, and Tianxing
  Chen.
\newblock {An Unexpected Robot Policy: Early Evaluations of GPT-6 Astra on
  RoboDojo and Beyond}.
\newblock \emph{arXiv preprint arXiv:2609.24170}, 2026{\natexlab{b}}.

\bibitem[Zhang et~al.(2026{\natexlab{c}})Zhang, Wang, Bauer, Yu, Huang, Wu, and
  Pavone]{robocop2026}
Xilun Zhang, Maggie Wang, Erik Bauer, Hong-Xing Yu, Huang Huang, Jiajun Wu, and
  Marco Pavone.
\newblock {Co-Evolving Robot Orchestrators and Policies through Deployment}.
\newblock \emph{arXiv preprint arXiv:2610.09228}, 2026{\natexlab{c}}.

\bibitem[Zhang et~al.(2026{\natexlab{d}})Zhang, Zhang, Gao, Li, Liu, Nie, Zhu,
  Qiu, Yan, Liu, Tang, Rao, Fang, Wei, Wang, Ding, and Yu]{harnessvla2026}
Yixian Zhang, Huanming Zhang, Feng Gao, Xiao Li, Zhihao Liu, Yi Nie, Chunyang
  Zhu, Jiaxing Qiu, Yuchen Yan, Jiyuan Liu, Wenhao Tang, Jiaji Rao, Zhengru
  Fang, Changxu Wei, Yu Wang, Wenbo Ding, and Chao Yu.
\newblock {Harness VLA: Steering Frozen VLAs into Reliable Manipulation
  Primitives via Memory-Guided Agents}.
\newblock \emph{arXiv preprint arXiv:2607.08448}, 2026{\natexlab{d}}.

\bibitem[Zhao et~al.(2026)Zhao, Huang, Mao, Liu, Kochenderfer, and
  Wong]{zhao2026retriever}
Linfeng Zhao, Haojie Huang, Jiayuan Mao, Weiyu Liu, Mykel Kochenderfer, and
  Lawson L.~S. Wong.
\newblock {Retriever: Composing the Perception-Reasoning-Action Loop for
  Long-Horizon Manipulation}.
\newblock \emph{arXiv preprint arXiv:2607.17213}, 2026.

\end{thebibliography}
